\documentclass[10pt]{article} 
\usepackage[accepted]{tmlr}

\usepackage{amsmath,amsfonts,bm}

\def\eqref#1{equation~\ref{#1}}

\def\1{\bm{1}}

\DeclareMathAlphabet{\mathsfit}{\encodingdefault}{\sfdefault}{m}{sl}
\SetMathAlphabet{\mathsfit}{bold}{\encodingdefault}{\sfdefault}{bx}{n}

\DeclareMathOperator*{\argmax}{arg\,max}

\usepackage{hyperref}
\usepackage{url}
\usepackage{soul}
\usepackage{lineno}

\usepackage{multirow}
\usepackage{array}
\usepackage{graphicx}
\usepackage{booktabs}
\usepackage{xspace}
\usepackage{enumitem}
\usepackage{subcaption} 
\usepackage{appendix}

\usepackage{array,booktabs}
\newcolumntype{C}[1]{>{\centering\arraybackslash}p{#1}}
\soulregister\ref7
\soulregister\cite7
\soulregister\ensuremath7

\usepackage{xcolor}

\usepackage{wrapfig}
\newcommand{\VLM}{LVLM\xspace} 
\newcommand{\ours}{PeKit\xspace}
\newcommand{\yollava}{Yo’LLaVA\xspace}

\newcommand{\thisismy}{This-Is-My-Img\xspace}
\newcommand{\thisismyv}{This-Is-My-Video\xspace}
\newcommand{\myparagraph}[1]{\noindent\textbf{#1}}

\definecolor{CR1}{rgb}{0.1, 0.80, 0.50} 
\definecolor{CR2}{rgb}{0.4, 0.18, 0.78} 

\newcommand{\be}{\mathbf{e}}

\usepackage{amsmath}
\DeclareMathOperator{\similarity}{sim}
\DeclareMathOperator{\AvgPool}{AvgPool}

\title{Personalization Toolkit: Training Free Personalization of Large Vision Language Models}

\author{Soroush Seifi\thanks{Providing contracted services at Toyota Motor Europe.} \quad
Vaggelis Dorovatas\textsuperscript{*} \quad
Matteo Cassinelli\textsuperscript{*} \quad
Fabien Despinoy \quad
Daniel Olmeda Reino \quad
Rahaf Aljundi\\[4pt]
\addr Toyota Motor Europe \\
}

\def\month{04}  
\def\year{2026} 
\def\openreview{\url{https://openreview.net/forum?id=5mbn3B0O29}} 

\begin{document}

\maketitle

\begin{abstract}

Personalization of Large Vision-Language Models (LVLMs) involves customizing models to recognize specific users or object instances and to generate contextually tailored responses. Existing approaches rely on time-consuming training for each item, making them impractical for real-world deployment, as reflected in current personalization benchmarks limited to object-centric single-concept evaluations.
In this paper, we present a novel training-free approach to LVLM personalization called \ours. We introduce a comprehensive, real-world benchmark designed to rigorously evaluate various aspects of the personalization task. \ours leverages pre-trained vision foundation models to extract distinctive features, applies retrieval-augmented generation (RAG) techniques to identify instances within visual inputs, and employs visual prompting strategies to guide model outputs. Our model-agnostic vision toolkit enables efficient and flexible multi-concept personalization across both images and videos, without any additional training. We achieve state-of-the-art results, surpassing existing training-based methods.

\end{abstract}
\section{Introduction}
\label{sec:intro}
\begin{figure*}[h]
    \centering    \includegraphics[width=\textwidth]{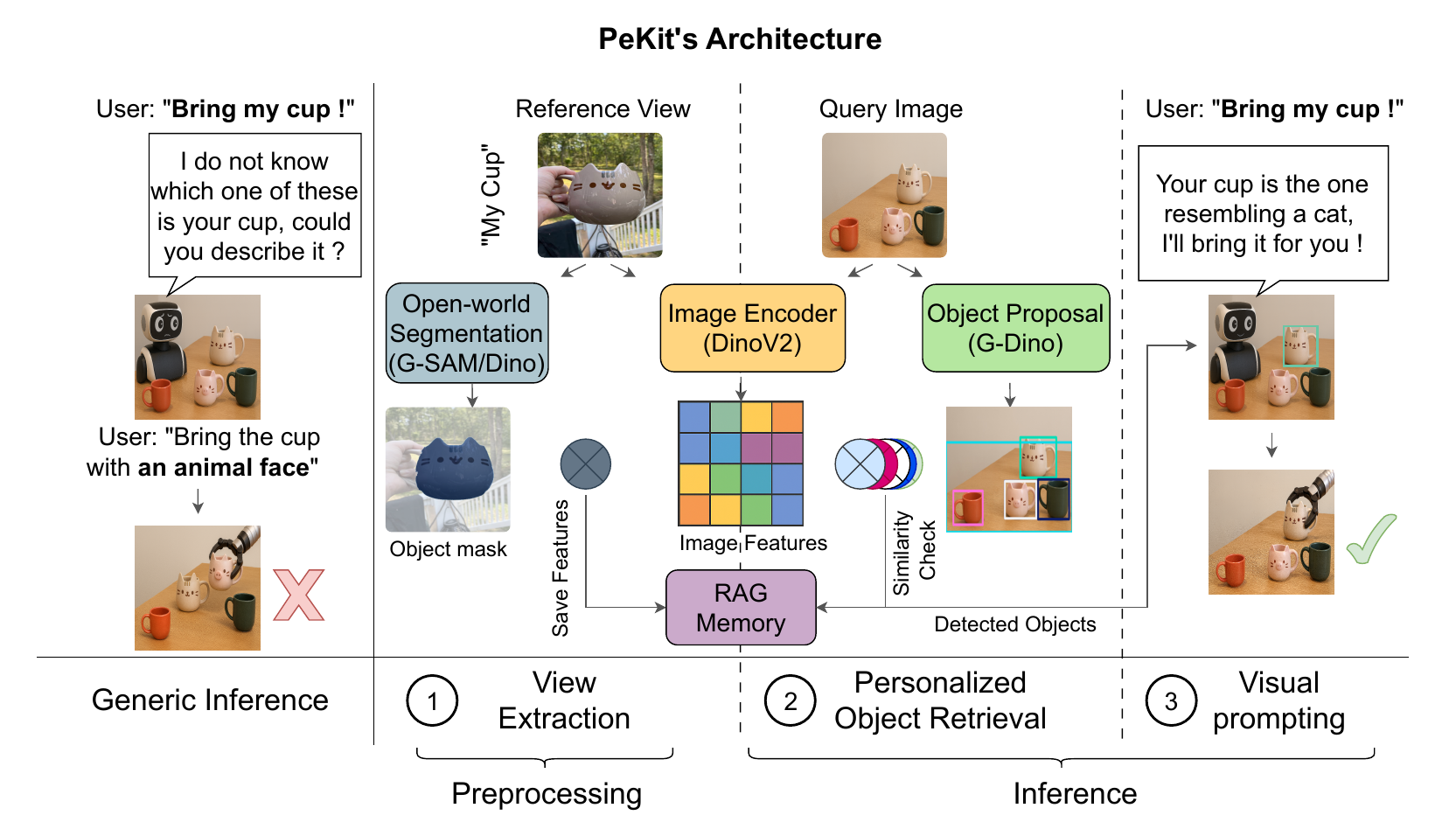}
    \caption{Illustration of the personalization task and \ours.
    Left: Without personalization, VLMs often fail to resolve named object references, leading to ambiguous responses.
    Right: \ours personalizes VLMs by (1) extracting patch-level features from a reference image into a RAG memory, (2) matching them with object proposals in the query image to retrieve the target object, and (3) guiding the VLM via visual prompting using the detected object’s bounding box and name.
    \ours is VLM-agnostic, supports multi-concept personalization, handles video input, and achieves state-of-the-art (SOTA) performance, enabling capabilities often unsupported by prior methods.}
    \label{fig:teaser}
\end{figure*}

Large Vision Language Models (LVLMs)~\cite{liu2024improved,liu2024visual,chen2024internvl,zhu2023minigpt,li2023blip,agrawal2024pixtral12b,wang2024qwen2} have demonstrated impressive capabilities in reasoning about visual content and answering visual questions across various domains. This suggests a great potential for deployment as visual assistants that can support users in their daily lives. However, current LVLMs are designed to provide generic, user-independent responses and recognize objects at the category level (Fig.~\ref{fig:teaser}, left). 



The task of personalizing vision-language models was introduced by~\cite{alaluf2024myvlm} to enable LVLMs to recognize specific object instances and answer relevant questions accordingly.  Existing approaches~\cite{alaluf2024myvlm,nguyen2024yo} rely on training for a specific personalized object, diverting the LVLM from its original capabilities and incurring a large computational cost. Follow-up approaches attempted at replacing test-time training with large scale pretraining for the personalization task~\cite{pham2024personalizedlargevisionlanguagemodels, pi2024personalized}, however, neither their effectiveness nor their scalability to various backbone models has been clearly demonstrated.


In this work, we argue that LVLM personalization can be approached without retraining the model parameters. We introduce a training-free approach that builds upon the strengths of pre-trained vision foundation models, the emerging capabilities of LLMs with in-context learning~\cite{dong2022surveyincontext} and retrieval-augmented generation (RAG)~\cite{lewis2020retrieval}. Our method localizes instances using open-world object detectors~\cite{liu2023grounding, kirillov2023segment, oquab2023dinov2} and stores reference instance-level features in memory banks, alongside their name and context. During inference, our retrieval module queries the memory bank and visual-prompts the LVLM. 
An overview of our approach is shown in Fig.~\ref{fig:teaser}.
With regards to the evaluation of personalization methodologies, existing  benchmarks primarily focus on object-centric, single-concept tasks with the personalized instance prominently in the image, falling short of capturing the complexity of real-world applications where an AI assistant needs to understand multiple users and their belongings in dynamic scenes and environments. To address this gap, we introduce a novel and challenging benchmark derived from a video personalization dataset~\cite{yeh2023meta}. Our benchmark not only features challenging single concept tasks but complex multi-concept interactions in addition to video question answering. 
We employ this benchmark to rigorously evaluate our method showing its effectiveness across diverse scenarios.

The key contributions of our work are:
1) We show that LVLM personalization is possible without training, enabling fast deployment.
2) We present a flexible method that supports multi-concept and video personalization across LVLMs using vision foundation models, RAG, and visual prompting.
3) We introduce a challenging benchmark that exposes current limitations and guides future research.\footnote{Our benchmark and code will be available on Github.}
4) We achieve state-of-the-art performance on various tasks across both existing datasets and our proposed benchmark, consistently outperforming previous methods.

We discuss the related work in Section~\ref{sec:related_work} followed by an introduction to our vision toolkit for LVLMs personalization in Section~\ref{sec:approach}. We then present our real-world benchmark and evaluate our approach in Section~\ref{sec:experiments}, and conclude in Section~\ref{sec:conclusion}.

\section{Related Work}
\label{sec:related_work}
\myparagraph{Text-to-Image Personalization.}
Personalizing text-to-image generation—i.e., generating images of a specific entity in novel contexts given a reference view—has been extensively explored. Early methods such as Textual Inversion~\cite{gal2022image}, DreamBooth~\cite{ruiz2023dreambooth}, and HyperDreamBooth~\cite{ruiz2024hyperdreambooth} achieve personalization by fine-tuning diffusion models for each entity, which limits scalability. More recent approaches, including InstantBooth~\cite{shi2024instantbooth}, JeDi~\cite{zeng2024jedi}, and Imagine~\cite{he2024imagine}, circumvent this issue by pretraining for personalization, thereby eliminating the need for test-time fine-tuning.

Beyond generation, several works have explored personalized image retrieval using CLIP~\cite{radford2021clip} as the backbone. For example, PALAVRA~\cite{cohen2022my} learns new concept tokens from a few user-provided images while keeping CLIP frozen, SEARLE~\cite{baldrati2023zero} maps reference images to pseudo-word tokens via a lightweight network and combines them with textual descriptions of desired changes for the Zero-Shot Composed Image Retrieval (ZS-CIR) task, and ConCon-Chi~\cite{rosasco2024concon} fuses image and text features through a learned composition network. In all these cases, retrieval is performed based on text–image similarity in CLIP space.

While these approaches focus on either text-to-image generation or retrieval, our work addresses a related but distinct challenge: identifying the same personalized object across different images and enabling personalized interactions with those objects through Large Vision–Language Models (LVLMs) such as LLaVA~\cite{liu2024visual}. We further posit that, unlike text-to-image personalization approaches that typically require training, personalization within LVLMs can be achieved without any adaptation of the underlying large language model.

\myparagraph{Large Vision–Language Model Personalization.}
Personalizing LVLMs was first introduced in MyVLM~\cite{alaluf2024myvlm}, which trains a concept head for specific objects on top of the CLIP \texttt{CLS} token. Similar to DreamBooth~\cite{ruiz2023dreambooth}, MyVLM uses rare tokens to encode personalized concepts, but this can introduce unintended behavior in language assistants and requires optimizing the LLM captioning loss for personalized conversations. Yo'LLaVA~\cite{nguyen2024yo} improves upon MyVLM by adding a dedicated token to the LLM head for each personalized object, learning concept tokens to describe them. However, this creates a challenging incremental classification problem~\cite{de2021continual}, and like MyVLM, it requires test-time training for each new concept, limiting scalability to one concept at a time.

To avoid test-time training, recent works leverage large-scale pretraining. PVIT~\cite{pi2024personalized} fine-tunes on synthetic personalized dialogues and uses reference images during inference. PLVLM~\cite{pham2024personalizedlargevisionlanguagemodels} aligns CLIP \texttt{CLS} and DINOv2~\cite{oquab2023dinov2} embeddings with the LLM. Both works, primarily target individuals and do not support object-level personalization. Besides, personalization remains query-specific and is largely limited to VQA tasks. 

In contrast, our approach introduces a modular, training-free framework that requires no retraining, scales naturally to multi-concept and video scenarios, and avoids relying on pre-trained tokens or reference images at inference time.

Finally, a concurrent work—Training-Free Personalization via Retrieval and Reasoning on Fingerprints~\cite{das2025training}—also proposes a training-free framework. It derives textual and visual fingerprint attributes from reference images and employs multi-step reasoning to recognize personal concepts. While effective, this method depends on a complex, hand-engineered pipeline involving attribute extraction, chain-of-thought reasoning, cross-modal verification, and pairwise comparisons, making it vulnerable to hallucinated or noisy textual attributes generated by the underlying LVLM. By contrast, our approach uses a simpler patch-level matching strategy that avoids reliance on potentially hallucination-prone textual descriptions.

\myparagraph{Visual Prompting.} It represents the usage of visual cues such as bounding boxes or arrows to guide Vision-Language Models. CLIP~\cite{radford2021clip} interprets these marks to modify its \texttt{CLS} token embedding accordingly~\cite{shtedritski2023does}. Set~of~Mark~Prompting~\cite{yang2023set} integrates GPT-4V with visual prompts using tools like MaskDINO~\cite{li2022mask}, SAM~\cite{kirillov2023segment}, Semantic SAM~\cite{li2023semantic}, and SEEM~\cite{zou2024seem}. ViPLLaVA~\cite{cai2024vip} enhances LLaVA~\cite{liu2024visual} to follow visual prompts by tuning on GPT-4V-labeled data. Contrastive Region Grounding (CRG)~\cite{Wan2024CRG} improves LLaVA’s focus on objects by contrasting token probabilities with and without target object masking. Our experiments show that LLaVA and other LVLMs can describe objects accurately with proper instruction and context. Training-free methods like CRG~\cite{Wan2024CRG} can further enhance attention if needed.

\section{Approach}
\label{sec:approach}
 This section outlines our personalization toolkit, coined as \textit{PeKit}, for enabling any LVLM to perform personalized detection and answer generation. We employ a three-stage pipeline: \textbf{View Extraction} to extract robust object-level features from reference images and store them in a memory module, \textbf{Personalized Objects Retrieval} to identify objects in the query image, and \textbf{Personalized Answer Generation} using visual prompting to generate user tailored and contextualized responses. We refer to Fig.~\ref{fig:teaser} for an illustration of our approach.
\subsection{Preliminary}
We consider a given \VLM, a large language model with visual understanding capabilities and  a set $P$ of all personalized objects introduced to the  \VLM. Each object $p\in P$ is associated with a set of reference images $\{I_p\}$, a name or identifier $n_p$ and optionally $c_p$ a context of the object.
Our objective is to generate a personalized response for all images containing $p$ during inference, while producing a general caption for any other image that does not contain any of the personalized objects.
The \VLM, e.g., LLaVA~\cite{liu2024visual} typically takes as input an input image $I_p$, a text query $Q$ and additional text as context or instruction.
\subsection{Training-free View Extraction}\label{sec:extraction}
\label{sec:training_views}
Existing \VLM personalization techniques depend on image-level representations of the objects’ training views~\cite{alaluf2024myvlm,nguyen2024yo}, which can lead to overfitting to the background of each object in the reference images, particularly for training-based approaches. To avoid such a bias,
our method first localizes the object in the image and extracts only its corresponding features. 
We utilize an open-vocabulary segmentation network $F_\text{ext}$ to extract object-level masks $S_p$ based on each object's generic category $k_p$ which can be deduced from the name or the context\footnote{Although using the semantic category $k_p$
 is recommended for optimal performance, our method attains state-of-the-art results even without it, relying solely on the generic category `main'. See section~\ref{sec:category} for further discussion.} 
\begin{equation}~\label{eq:ext}
    S_p = F_\text{ext}(I_p,k_p).
\end{equation}
We construct the average embedding vector $\be_p$ for object $p$ by average pooling of the embedding vectors produced by the image encoder $F_\text{emb}$ on the image $I_p$ over the region defined by the object-level mask $S_p$:
\begin{equation}
    \be_p=\AvgPool(F_\text{emb}{(I_p),S_p)}\in\mathbb{R}^{D_h}.
\end{equation}
Considering $N$ reference images, we concatenate all object embedding vectors $\be_p^i$ (of the object $p$ pooled over the $i$-th reference image $I_p^i$) into a matrix $E_p = \left[\be_p^1, \dots, \be_p^N\right]\in\mathbb{R}^{D_h \times N}$.

\myparagraph{Memory module.}  After extracting the personalized objects' embeddings, we store  each object's relevant properties in our  memory module.  The memory module is represented by a set $\mathcal{M}$ of object-specific entries:
\begin{equation}
    \mathcal{M}=\{ (E_p, (n_p, c_p)) \}_{p\in P},
\end{equation}
where $n_p$ is the identifier or the name of the personalized object $p$, and $c_p$ is the context of the object, which can contain prior knowledge such as characteristics, background story or even relation to other personalized objects.
When the number of personalized objects scales, the memory module $\mathcal{M}$ is easily converted into a Vector database, where nearest neighbor approximate search is deployed to retrieve instances matching a given query~\cite{han2023vector} ensuring efficiency and scalability.
\subsection{Personalized Object Retrieval}
\label{sec:retrieval}
During inference, our goal is to determine whether a personalized object is present in the provided image $I$.
We use any available object proposal technique $F_\text{prop}$ to generate a set of proposals (e.g., bounding boxes) $O=\{o_i\}=F_\text{prop}(I)$ for potential object occurrences within the image $I$.
Then for each proposal $o_i$, we calculate its object-level average embedding vector:
\begin{equation}
\be_{o_i} = \AvgPool(F_\text{emb}{(I),o_i}).
\end{equation}
We define the retrieval module $\mathcal{R}$ that takes an object embedding vector $\be_{o_i}$ and retrieves the memory entry $(E_j, (n_j, c_j))$ for a matching object $j$ as:
\begin{equation}
\mathcal{R}(\be_{o_i}) = 
\begin{cases} 
    (E_j, (n_j, c_j)),\\
    \quad \text{if } j = \argmax_l\left\{\similarity(E_l, \be_{o_i})\right\} \\
     \quad \text{and } \similarity(E_j,\be_{o_i}) > \tau \\
    \phi,\, \text{otherwise}
\end{cases}
\end{equation}
where $\similarity(E_l,\be_{o_i})= \argmax_k\left\{\similarity(\be_l^k, \be_{o_i})\right\}$.
We calculate the proposal's similarity to the embeddings of the training views for all personalized objects.
Any similarity measure (e.g., cosine similarity) can be employed for this purpose.
We set a constant threshold $\tau$ to identify the personalized objects.
We discard the object proposals in which no matching object is found by the retrieval module $\mathcal{R}$.
 Our method inherently supports the detection of multiple personalized objects.
\subsection{Personalized Answer Generation}
\label{sec:vprompting}
Once a personalized object is identified, our method generates captions specifically about that object, distinct from general captions a standard \VLM would produce. This involves emphasizing the detected object and incorporating prior knowledge about it. We achieve this through visual prompting by overlaying bounding boxes on the image and querying the \VLM to generate captions or answer questions focused on these objects. We use distinctive colors to differentiate recognized objects. 
We provide the \VLM with the object identifier $n_j$ (e.g., the instance name) and possibly a context $c_j$ for each personalized object. The \VLM incorporates this context and responds to queries using the given name $n_j$. For multiple personalized objects, we instruct the \VLM for each bounding box, name, and context. 
We refer to Appendix C.1 for the exact prompt format. 
\subsection{Video-personalization}\label{vidpers}
To conduct personalized video question answering (VQA), we apply our  method to the video frames.  We first identify any presence of a personalized object, then each personalized object is consistently visually prompted using the same bounding box color across all frames to maintain identity. The video frames---including those with visual prompts and those without any detected personalized objects---are  fed into a video large language model.

The model is then prompted to generate captions or answer questions using each detected object's identifier \( n_j \), an optional context \( c_j \), and the unique bounding box color used for the visual prompting of that object instance (Appendix C.1). 
\subsection{Choice of Vision Tools}
We deploy  GroundedSAM~\cite{ren2024grounded} as the segmentation network $F_\text{ext}$ and ablate using GroundingDINO~\cite{liu2023grounding}, where the mask is represented by the object’s bounding box. We use DINOv2~\cite{oquab2023dinov2} as the image encoder $F_\text{emb}$ to extract patch-level features of the objects and ablate using CLIP~\cite{radford2021clip} in Appendix A.1. 
During inference, GroundingDINO~\cite{liu2023grounding} is queried with the term `object' to generate object proposals.
\section{Experiments}
\begin{figure*}[t]
    \centering
    \includegraphics[width=\textwidth]{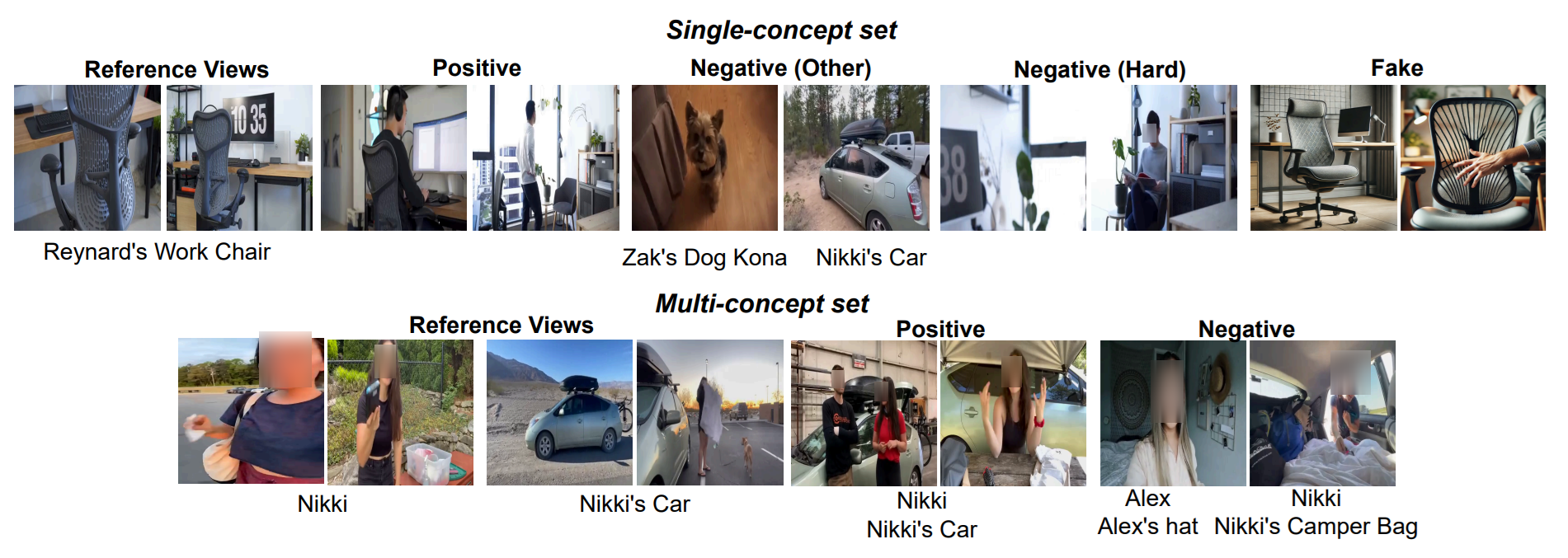}
    \caption{Our proposed evaluation set \thisismy, built on the This-Is-My dataset~\cite{yeh2023meta}: Example reference views and validation samples from the single-concept category `Reynard's Work Chair' and the multi-concept category `Nikki' and `Nikki's Car'.
}
    \label{fig:this_is}
\end{figure*}
\label{sec:experiments}
In this section we present the considered benchmarks and  introduce our LVLM personalization benchmark.  We then compare our approach with  SOTA  personalization methods and baselines under various settings.
\subsection{Existing Benchmarks}
We consider the datasets from \textbf{Yo’LLaVA}~\cite{nguyen2024yo} and \textbf{MyVLM}~\cite{alaluf2024myvlm}. Yo’LLaVA includes 40 categories of objects, buildings, and people, with 4--10 images per category for training and validation. It also features a VQA benchmark with multiple-choice questions (A/B). MyVLM comprises 29 object categories, each with 7--17 images, using 4 images for training and the rest for validation, with final accuracies averaged over 5 runs. For both datasets, we use only reference images and the semantic category of each object for training view extraction (see \ref{sec:training_views}), discarding all other data, such as the ground-truth captions and negative images.
\subsection{Real-world Personalization Benchmark}
\textbf{Yo’LLaVA}~\cite{nguyen2024yo} and \textbf{MyVLM}~\cite{alaluf2024myvlm} focus mainly on object-centric scenes under controlled conditions, lacking real-world complexity such as background clutter, occlusion, and diverse contexts—limiting their suitability for evaluating robust LVLM personalization.

We introduce the \textbf{\thisismy} benchmark, derived from the \textit{This-Is-My} dataset~\cite{yeh2023meta}, originally designed for video-level detection of personalized objects in realistic environments. Our benchmark includes several splits targeting specific personalization capabilities.

\noindent\textbf{Reference Views:} Five reference frames per object extracted from training segments. Unlike prior datasets, objects here may be partially visible, distant, or poorly lit (see Fig.~5, Appendix).

\noindent\textbf{Single-Concept Eval Set:} Set of images designed to evaluate a single personalized object. Fig.~\ref{fig:this_is} shows examples. For each personalized object, images are labeled as:
\begin{itemize}[leftmargin=*]
\item \textit{Positive}: Includes only images where the personalized object appears and is used to test how well the method can correctly detect all occurrences of that object, that is, recall or \textit{ positive accuracy} in \textbf{Yo’LLaVA}~\cite{nguyen2024yo} and \textbf{MyVLM}~\cite{alaluf2024myvlm}.

\item \textit{Negative (Hard)}: Images taken from video segments of the personalized object \textbf{where the object is not visible}. This tests method's overfitting to the background associated with the personalized object in the reference views.

\item \textit{Negative (Other)}: Images sourced from video segments of other personalized objects. This set measures method's robustness to intra-instance and the general dataset bias.

\item \textit{Negative (Fake)}: AI-generated images (via GPT-4o) that mimic the personalized object in different perspectives and environments while making sure the specific characteristics of the fake object are different from the real personalized object. These samples are used to assess the robustness of the methods against high visual similarity combined with a distribution shift.

\item \textit{Single-Concept VQA:} Multiple choice questions based on positive images, following the Yo'LLaVA benchmark VQA format.

\end{itemize}

\noindent\textbf{Multi-Concept Set:} This subset comprises images containing pairs of personalized concepts, such as person–object pairs or person–person pairs, fig \ref{fig:this_is}.
We extend the personalized instances from the original This-Is-My dataset with additional instances that frequently co-occur with them. For each pair, we collect all images containing both personalized concepts and partition them into two subsets.
Positive samples include images in which both concepts appear simultaneously, whereas negative samples comprise images from all other pairs.
Illustrative examples are presented in Fig.~\ref{fig:this_is}.

\begin{itemize}[leftmargin=*]
\item \textit{Multi-Concept VQA:} This component provides open-ended questions associated with each image in the multi-concept set, addressing both personalized concepts depicted. The set includes 55 images and their corresponding QA pairs.
\end{itemize}

\noindent\textbf{Video-QA Set:} Original video clips from This-Is-My dataset are coupled with open-ended questions regarding the personalized object in each video, designed to evaluate both temporal reasoning and the method adaptability to video-based inputs. 

Overall, \thisismy offers a comprehensive and realistic benchmark for LVLM personalization, requiring accurate recognition and reasoning over personalized objects in cluttered and dynamic conditions. See Appendix B for more details on our proposed benchmark. 

\subsection{Metrics and Evaluation Setting}
\myparagraph{Single-Concept Metrics.} We adopt the metrics and notation from Yo’LLaVA~\cite{nguyen2024yo} and MyVLM~\cite{alaluf2024myvlm}. We report \textit{Positive Accuracy (Recall)}—the proportion of correctly identified positives, \textit{Negative Accuracy (Specificity)}—correctly identified negatives, and \textit{Weighted Accuracy}, their average across $n$ personalized objects. We also include \textit{Precision}, the ratio of true to predicted positives, averaged over $n$ objects, to capture the accuracy–precision trade-off.
For single-concept VQA, accuracy is the percentage of correctly answered multiple-choice questions. Following MyVLM, we additionally report \textit{CLIPScore} (caption–image similarity) and \textit{Personalization Recall}, the fraction of captions correctly mentioning the target concept.

\myparagraph{Multi-Concept Metrics.}
Positive accuracy measures images where both personalized objects are correctly detected; negative accuracy covers those where at least one is correctly \textbf{not} detected.

\myparagraph{Video and Multi-Concept VQA Metric.}
For open-ended VQA, we use an autograding protocol~\cite{maaz2023video} with ChatGPT-3.5 to assess contextual similarity between predicted and ground-truth answers (see Appendix C.2).

\myparagraph{Implementation Details.}
The modularity of our method makes it generic regarding the choice of the \VLM model. We use LLaVA1.5-13B~\cite{liu2024improved} as our primary \VLM model for consistency with previous works. We pair \ours with InternVL2-26B~\cite{chen2024internvl} , a state-of-the-art \VLM to ablate. We use LLaVA-OneVision-Qwen2-7B as our default video model and ablate with InternVideoChat2.5 as a superior video model.
We utilize Cosine Similarity with a constant threshold of $\tau$=0.75 for detecting personalized objects across all datasets (\ref{sec:retrieval}). Refer to  Appendix A.6 for further details on compute and memory overhead of our personalization toolkit.

\begin{table}[t]
\caption{Visual recognition performance (\%) on existing datasets. \ours achieves state of the art performance without training. Cited entries are taken from their respective papers.}
      \label{tab:recognition}
\centering
\begin{tabular}{lcccc}
\toprule
\multicolumn{5}{c}{\textbf{MyVLM Dataset}}\\
\cmidrule(lr){1-5}
\multirow{2}{*}{Method/Metric} & \multirow{2}{*}{Precision} & \multicolumn{3}{c}{Accuracy} \\
& & Positive & Negative & Weighted \\
\midrule
MyVLM \cite{alaluf2024myvlm}        & --              & 96.6              & 90.9              & 93.8 \\
Yo'LLaVA \cite{nguyen2024yo}       & --              & \underline{97.0}  & 95.7              & 96.4 \\
\ours (G\mbox{-}DINO)               & \underline{79.1} & 94.3             & \underline{98.8}  & \underline{96.5} \\
\ours (G\mbox{-}SAM)                & \textbf{82.3}    & \textbf{97.6}    & \textbf{99.0}       & \textbf{98.3} \\
\addlinespace[2pt]
\multicolumn{5}{c}{\textbf{Yo' LLaVA Dataset}}\\
\cmidrule(lr){1-5}
\multirow{2}{*}{Method/Metric} & \multirow{2}{*}{Precision} & \multicolumn{3}{c}{Accuracy} \\
& & Positive & Negative & Weighted \\
\midrule
Yo'LLaVA \cite{nguyen2024yo}       & --              & \textbf{94.9}    & 89.8              & 92.4 \\
\ours (G\mbox{-}DINO)               & \textbf{77}     & 89.9             & \textbf{98.9}     & \underline{94.4} \\
\ours (G\mbox{-}SAM)                & \underline{74.8}& \underline{91.0}   & \underline{98.7}  & \textbf{94.9} \\
\bottomrule
\end{tabular}
\end{table}


\begin{table}[t]
\centering
\caption{Visual recognition accuracy (\%) on the This-Is-My-Img dataset. MyVLM and \yollava produce many false positives in a challenging real-world scenario.}
\label{tab:this-is-my}
\begin{subtable}{\linewidth}
\caption{Single-concept results.}
\label{tab:single_concept}
\centering
\begin{tabular}{l r r r r r r}
\toprule
Method/Metric & Precision & \multicolumn{4}{c}{Accuracy} & Avg. \\
\cmidrule(lr){3-6}
& & Pos. & Other & Hard & Fake & \\
\midrule
MyVLM   &  8.0 & \textbf{88.1} & 14.6 &  4.7 & 54.2 & 33.9 \\
\yollava & 42.1 & 87.1 & 84.4 & 61.9 & \textbf{61.4} & 67.3 \\
Ours    & \textbf{90.1} & 69.0 & \textbf{99.9} & \textbf{96.0} & 59.3 & \textbf{82.8} \\
\bottomrule
\end{tabular}
\end{subtable}

\vspace{0.8em}

\begin{subtable}{\linewidth}
\caption{Multi-concept results.}
\label{tab:multi_concept}
\centering
\begin{tabular}{l r r r r}
\toprule
Method/Metric & Precision & \multicolumn{2}{c}{Accuracy} & Avg. \\
\cmidrule(lr){3-4}
& & Pos. & Neg. & \\
\midrule
\yollava & 10.0 & \textbf{83.6} & 25.0 & 54.3 \\
Ours    & \textbf{96.1} & 45.4 & \textbf{99.8} & \textbf{72.6} \\
\bottomrule
\end{tabular}
\end{subtable}

\end{table}

\subsection{Visual-Recognition }
\textbf{Previous Datasets} Tab.~\ref{tab:recognition} compares our method to MyVLM~\cite{alaluf2024myvlm} and Yo’LLaVA~\cite{nguyen2024yo} on their respective benchmarks. On MyVLM benchmark, \ours achieves a new SOTA, improving both positive and negative accuracy with an average gain of 1.9\%. On Yo’LLaVA benchmark, it increases negative accuracy by 8.9\%, with slightly lower positive accuracy, resulting in a  2.5\% improvement in weighted accuracy. While \ours may miss some difficult instances, its low false positive rate helps avoid incorrect personalized outputs, favoring generic captions when uncertain. We ablate the choice of an open-world object detector, G-DINO~\cite{liu2023grounding}, compared to the open-world semantic segmentation model, G-SAM~\cite{ren2024grounded}, for~$F_\text{ext}$ in Eq.~\ref{eq:ext}. The results show that our method outperforms previous methods in both cases. However, the more precise segmentation model, which extracts only patches of the object of interest, achieves the best accuracy on average.



\begin{table}[t]
  \centering
  \caption{Captioning metrics on MyVLM dataset. Higher is better.}
  \label{tab:myvlm_vqa}
  {
  \begin{tabular}{lcc}
    \toprule
    Method & CLIPScore & Personal. Recall \\
    \midrule
    MyVLM~\cite{alaluf2024myvlm} & \underline{27.6} & \underline{94.76} \\
    \ours~(LLaVA)                & \textbf{30.2}    & \textbf{97.1}     \\
    \bottomrule
  \end{tabular}
  }
\end{table}

\begin{table}[t]
  \caption{Single-concept VQA accuracy on Yo’LLaVA and \thisismy datasets.}
  \label{tab:vqa_yollava_single}
  \centering
  \begin{minipage}{0.48\linewidth}
    \centering
    {
    \begin{tabular}{lc}
      \toprule
      \multicolumn{2}{c}{\textbf{Yo’LLaVA}}\\
      \multicolumn{2}{c}{VQA Accuracy}\\
      \midrule
      LLaVA \cite{nguyen2024yo}        & 89.9 \\
      Yo'LLaVA \cite{nguyen2024yo}     & 92.9 \\
      \ours\ (LLaVA)                   & \underline{93.4} \\
      \ours\ (InternVL)                & \textbf{95.9} \\
      \bottomrule
    \end{tabular}
    }
  \end{minipage}
  \hfill
  \begin{minipage}{0.48\linewidth}
    \centering
    {
    \begin{tabular}{lc}
      \toprule
      \multicolumn{2}{c}{\textbf{\thisismy\ Single-Concept}}\\
      \multicolumn{2}{c}{VQA Accuracy}\\
      \midrule
      LLaVA             & 72.8 \\
      Yo'LLaVA          & 67.1 \\
      \ours\ (LLaVA)    & \underline{77.1} \\
      \ours\ (InternVL) & \textbf{84.2} \\
      \bottomrule
    \end{tabular}
    }
  \end{minipage}
\end{table}

\begin{table}[t]
  \centering
  \caption{VQA accuracy on \textbf{\thisismy\ Multi-Concept} and \textbf{\thisismyv} datasets.}
  \label{tab:vqa_multi_thisismyv}
    \begin{minipage}{0.48\linewidth}
  \small
    \centering
    {
    \begin{tabular}{lc}
      \toprule
      \multicolumn{2}{c}{\textbf{\thisismy\ Multi-Concept}}\\
      \multicolumn{2}{c}{VQA Accuracy}\\
      \midrule
      LLaVA             & 49.0 \\
      Yo'LLaVA          & 12.7 \\
      \ours\ (LLaVA)    & \underline{56.3} \\
      \ours\ (InternVL) & \textbf{63.6} \\
      \bottomrule
    \end{tabular}
    }
  \end{minipage}
  \hfill
  \begin{minipage}{0.48\linewidth}
  \small
    \centering
    {
    \begin{tabular}{lc}
      \toprule
      \multicolumn{2}{c}{\textbf{\thisismyv}}\\
      \multicolumn{2}{c}{VQA Accuracy}\\
      \midrule
      LLaVA-OneVision-Qwen2          & 23.0 \\
      \ours\ (LLaVA-OneVision-Qwen2) & 35.0 \\
      InternVideoChat2.5             & \underline{56.6} \\
      \ours\ (InternVideoChat2.5)    & \textbf{61.3} \\
      \bottomrule
    \end{tabular}
    }
  \end{minipage}
\end{table}


\begin{figure*}[t]
    \centering
\includegraphics[width=\linewidth]{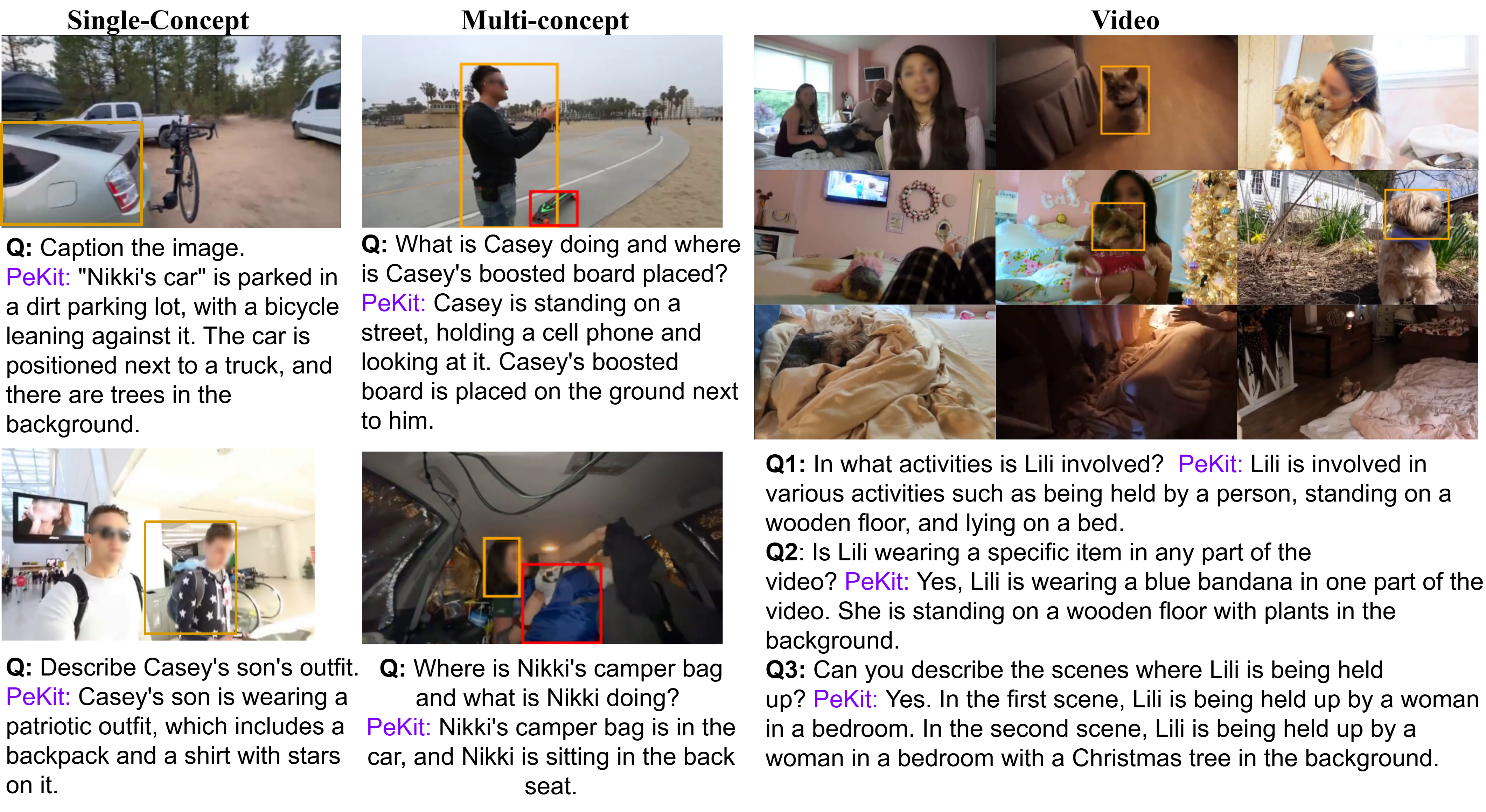}
        \caption{\textbf{Qualitative Results:} PeKit handles a range of personalization tasks, encompassing both single- and multi-concept personalization in images and videos. For video personalization, the VLM model can reliably track the target object across frames using only a few confidently annotated instances. One representative frame is shown per scene.}
    \label{fig:qualitative}
\end{figure*}
\textbf{\thisismy Benchmark.} Tab.~\ref{tab:this-is-my} summarizes benchmark performance. For single-concept images, \textbf{MyVLM} achieves high positive accuracy but poor negative accuracy (14.6\% on \textit{other}, 4.7\% on \textit{hard}), indicating a bias toward positive responses and reliance on scene context over object details. \textbf{\yollava} performs similarly but better (87.1\% positive, 84.8\% on \textit{other}, 61.9\% on \textit{hard}). In contrast, \textbf{\ours} attains much stronger negative accuracy (99.9\%, 96.0\%) and 90.1\% precision—48\% higher than \yollava—showing far fewer false positives.

On fake images, differences narrow. All models struggle with stylistically similar objects; image-level approaches like MyVLM and \yollava slightly better detect domain shifts. Nonetheless, \ours remains robust to domain variation despite some difficulty with near-identical objects, achieving the best overall balance with 82.86\% average accuracy, surpassing \yollava (67.38\%) and MyVLM (33.92\%).

In multi-concept settings, \ours shows slightly lower dual-object accuracy but far higher precision and negative accuracy, yielding an 18.3\% overall gain over \yollava. These results highlight our training-free method’s robustness and low false-positive rate, especially in complex, real-world scenarios.

Overall, this evaluation underscores the challenge of personalization beyond object-centric imagery, advancing toward realistic benchmarks for intelligent visual assistants.
\subsection{Visual-Question Answering}
Tabs.~\ref{tab:myvlm_vqa}, \ref{tab:vqa_yollava_single}, and \ref{tab:vqa_multi_thisismyv} evaluate \ours on personalized object VQA across multiple scenarios using the Yo’LLaVA~\cite{nguyen2024yo} and \thisismy benchmarks. For completeness, we also compare against MyVLM on its dataset using CLIPScore~\cite{hessel2021clipscore} and Personalization Recall, as it lacks a VQA split.

\ours consistently outperforms baselines on both single- and multi-concept VQA tasks. On the MyVLM dataset, PeKit surpasses MyVLM in generating image-aligned captions (CLIPScore) and correctly referencing personalized concepts (Personalization Recall). On the \yollava benchmark, \ours outperforms Yo’LLaVA without requiring any fine-tuning, special tokens, or model modifications.

On the \thisismy benchmark, Yo’LLaVA performs on par with the base LLaVA, while \ours yields a 5\% improvement, showing strong visual recognition and reasoning. In multi-concept cases, Yo’LLaVA degrades base model performance, whereas \ours delivers a consistent 7\% gain. Moreover, \ours generalizes effectively to the video domain (\ref{vidpers}), unlike prior approaches, which are not readily extendable to video. \ours improves the base model by 12\% and maintains gains when combined with stronger instruction-following backbones.


In summary, our experiments demonstrate that \ours consistently outperforms training-based personalization methods across diverse benchmarks, including single- and multi-concept VQA tasks. Its ability to generalize without fine-tuning, maintain performance in multi-concept scenarios, and extend effectively to video settings highlights its robustness and scalability for real-world personalized visual understanding.

\subsection{Qualitative Results}
Fig.~\ref{fig:qualitative} presents examples of  \ours  performing various tasks on the proposed benchmark, \thisismy. In the left column, \ours successfully identifies personalized single-concept objects, even in ambiguous scenarios involving multiple instances from the same object category. In the middle column, it demonstrates the ability to answer questions involving multi-concept pairs. Lastly, without any modifications, the model can be applied to video inputs by visually prompting the model with detected personalized objects in the video. Further qualitative comparisons with the base LLaVA model, MyVLM, and Yo’LLaVA are presented in Appendices D.1, D.2, and D.3.
\subsection{Ablation and Analysis}
A comprehensive analysis of our method is provided in  Appendix.A, where we analyze robustness under various settings. Specifically, we examine the effects of altering the backbone encoder \textit{$F_{emb}$}, the number of reference images per object (\textit{N}), the number of personalized objects (\textit{P}) in the dataset, the detection threshold ($\tau$) and the name ($n_p$) of the personalized objects.

In the following subsection, we additionally examine the robustness of our approach with respect to a requirement introduced by our framework, namely the semantic category $k_p$ assigned to each object (see Section~\ref{sec:extraction}).

\subsubsection{Semantic Category $k_p$}\label{sec:category}
As described in Section~\ref{sec:extraction}, we use the semantic category $k_p$ of each object in the reference image $I_p$ to derive its corresponding object-level mask $S_p$.

We note that, in practice, introducing an object for personalization typically involves specifying its name. In many cases, the name itself implicitly encodes the semantic category—e.g., `My book' or `Jack's car'—which can then be leveraged by an open-vocabulary segmentation model for view extraction.

Therefore, for the MyVLM dataset, where the object names are generic, we directly input them into our open-vocabulary detector. Similarly, on \thisismy dataset, aside from the people in the dataset (queried with the category `Man/Woman'), all other object names in this dataset include their semantic category (e.g., Alex's hat, See Appendix.B).

For the Yo'LLaVA dataset, some concept names such as Vietnamese individual names do not directly indicate the semantic categories. We compare our method's performance on the Yo'LLaVA dataset by providing the open-vocabulary segmentation model with semantic categories, dataset-provided names, or the generic term `main' during reference view extraction. As shown in Table~\ref{tab:gsam_vocab}, PeKit achieves state-of-the-art performance even without relying on semantic categories of personalized objects. Note that during inference we consistently use the term `object' to extract object proposals using G-Dino.

Consequently, when the reference image for a personalized object is representative, with the subject clearly dominating the scene, our method remains competitive even without an explicitly provided semantic category $k_p$. If needed, a lightweight open-world classifier can be employed to infer this label prior to processing the object within our pipeline.
\begin{table}[h]
 \centering
   \caption{Reference view vocabulary ablation on Yo'LLaVA dataset. \ours achieves SOTA results even using the generic category `main' to extract the personalized objects from the reference views.}
\resizebox{0.75\textwidth}{!}{
\begin{tabular}{|c|c|c|c|c|}
\hline
Method/Metric & Precision & Positive Acc & Negative Acc & Weighted Acc \\
\hline
Yo' LLaVA & -  & \textbf{94.9} & 89.8 & 92.4 \\
\ours (categories) & \textbf{74.8} & \underline{91} & \underline{98.7} & \textbf{94.9} \\
\hline
\ours (`main') & 72.9 & 86.9 &  \textbf{98.9} & 92.9 \\
\ours (Vietnamese names) & \underline{73.8} & 88 & \textbf{98.9} & \underline{93.45} \\
\hline
\end{tabular}
}

  \label{tab:gsam_vocab}
\end{table}




\section{Limitations and Future Work}
\subsection{Noisy Reference Views}
\begin{figure*}[h]
\centering
\includegraphics[width=0.65\linewidth]{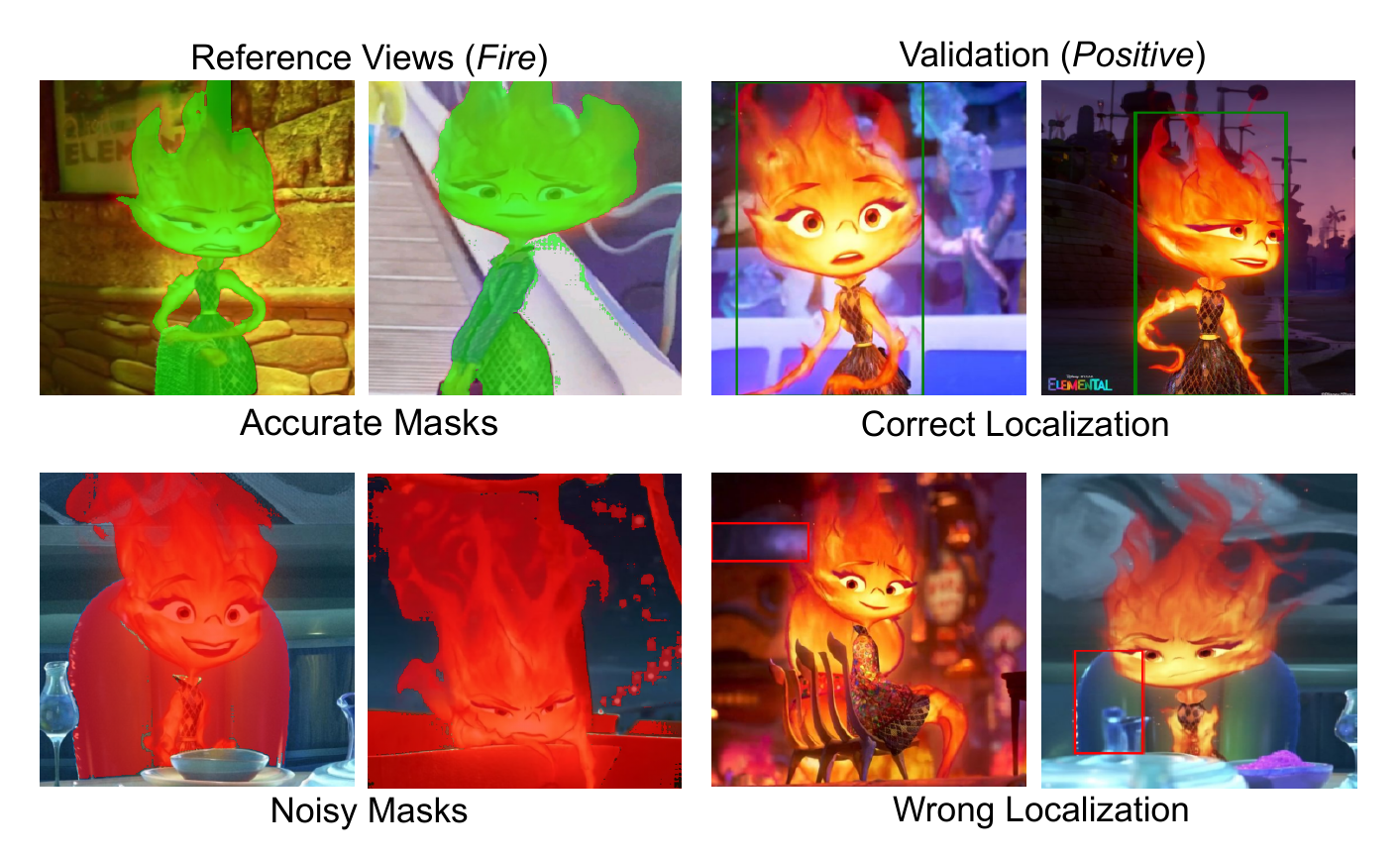}
\caption{Noisy reference views: Poor segmentation masks may affect the visual prompting stage and degrade PeKit's performance.}
\label{fig:failure_masks}
\end{figure*}

Our method relies on instance masks generated by a segmentation network. Inaccurate or noisy masks can lead to false positives during inference. As shown in Figure~\ref{fig:failure_masks}, poor masks (highlighted in red) result in incorrect matches (red bounding boxes). Applying clustering to reference features may help filter out erroneous views and improve matching reliability.
\subsection{Small Object Representation}
\begin{figure*}[h]
\centering
\includegraphics[width=0.65\linewidth]{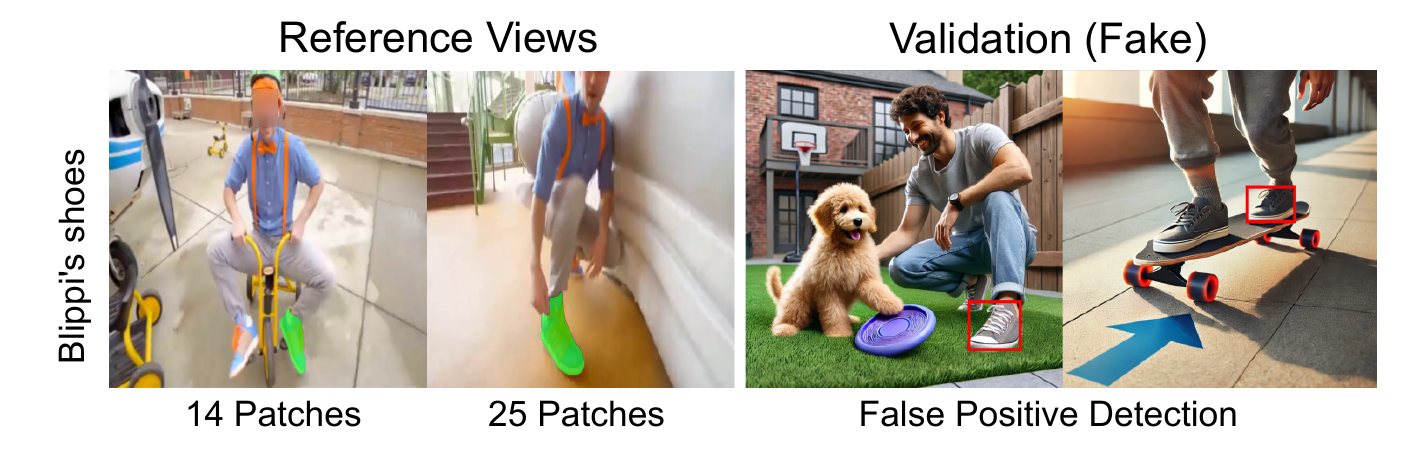}
\caption{Small reference objects: Due to DINO’s fixed patch size of $14 \times 14$, the resulting embeddings are relatively coarse, which can lead to increased false positives, especially for small or fine-grained objects.}
\label{fig:failure_small}
\end{figure*}

Due to the stride factor (=14) in the DINOv2 encoder, feature extraction occurs at a lower resolution than the original image. This limits detail capture for small objects, as illustrated in Figure~\ref{fig:failure_small}, where only coarse features (e.g., color or category) are encoded, increasing false positives. A potential solution is to crop and resize small objects to DINOv2’s native resolution ($518\times518$) before embedding.
\subsection{Contextual Object Relationships}
Our instance-level detection approach could benefit from contextual cues. For example, identifying \textit{Alex’s everyday bag} is more reliable when \textit{Alex} is present in the scene. Future work will explore extracting object relationships from reference images and integrating them into the \VLM as prior knowledge.

\section{Discussion and Conclusion}
In this work, we introduced PeKit, a training-free, plug-and-play toolkit for LVLM personalization that combines vision foundation models with retrieval-augmented generation and visual prompting. Our approach outperforms existing training-based methods without requiring any fine-tuning or additional data beyond the personalized concept inputs.
To evaluate our method, we proposed a challenging new benchmark that reflects more realistic scenarios involving multi-concept and video personalization, significantly increasing the complexity of the visual recognition task. PeKit demonstrates strong robustness and scalability across both multi-concept and video settings.
Our benchmark reveals strong performance and shows some improvement points, particularly in handling complex temporal and multi-concept reasoning. We believe that PeKit establishes a new efficient training-free baseline for future research in LVLM personalization.
\label{sec:conclusion}
\section{Ethical Considerations}
\ours is a training-free personalization approach for LVLMs designed to assist users in everyday tasks. It enables local and efficient customization without the need for retraining, supporting applications such as personalized robotics and visual assistants (see Figure~\ref{fig:teaser} and Figure 10 Appendix).

We recognize potential risks if the method were misused for surveillance or applied without user consent. To address these concerns, \ours operates entirely on-device, never transmits data externally, and works on extracted features rather than storing raw images. Personalization remains fully user-controlled, requiring explicit input.

\bibliography{main}
\bibliographystyle{tmlr}

\clearpage
\appendix
\appendixpage
\addappheadtotoc
\begin{appendices}
\section{Ablation}
In this section we ablate various aspects of our personalization method primarily using the \yollava dataset as it serves as a well-established benchmark.
\subsection{Retrieval Module Backbone}
We present an ablation study of the image encoder $F_{emb}$, which serves as the backbone of the Retrieval Module and is responsible for extracting object features and performing matching. Table~\ref{tab:supp_retrieval_abl} shows a noticeable gap between CLIP and DINO, indicating that DINO's visual features are more discriminative and better suited for retrieval tasks. Between the two DINO variants (base and large), the larger version delivers slightly better performance, the difference is marginal indicating that any DINOv2 backbone can be deployed.

 \begin{table}[h!]
 \centering
    \caption{Feature extractor ablation. DINO features significantly outperform CLIP features.}
    \resizebox{0.6\linewidth}{!}{\small
\begin{tabular}{|c|c|c|c|c|}
\hline
\multicolumn{5}{|c|}{\textbf{Yo' LLaVA Dataset}} \\
\hline
\multirow{2}{*}{Method/Metric} & \multicolumn{1}{c|}{Precision} & \multicolumn{3}{c|}{Recall} \\
 & & Positive & Negative & Weighted \\
\hline
DINOv2 (Large) & \underline{74.8} & \textbf{91} & \textbf{98.7} & \textbf{94.9}\\
\hline
DINOv2 (Base) & \textbf{75.5} & \underline{90.8} & \textbf{98.7} & \underline{94.7}\\
\hline
CLIP (Large) & 69  & 78.3 & 98.2 & 88.3\\
\hline
\end{tabular}
}
    \label{tab:supp_retrieval_abl}
\end{table}
\subsection{Number of Reference Images (\textit{N})}
\begin{figure}[h]
    \centering
    \includegraphics[width=0.6\linewidth]{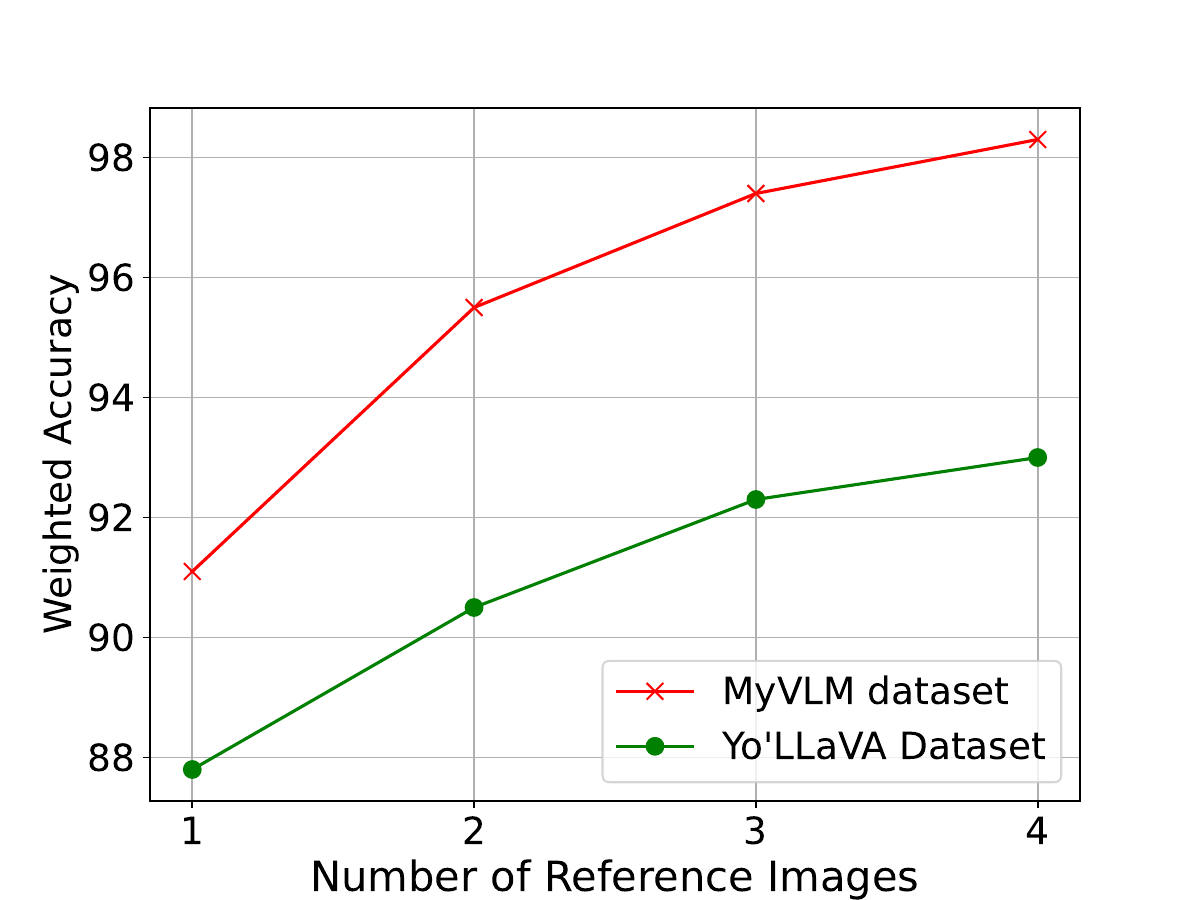}
    \caption{Ablation on N: Average weighted visual recognition accuracy as a function of number of reference images. Increasing the number of reference images improves performance, but \ours is robust with just one reference image.} 
    \label{fig:number_ref}
\end{figure}

Since our approach does not require a training phase, a key question is how many reference images are needed for robust visual recognition of personalized objects. Figure~\ref{fig:number_ref} shows that our method performs well with just one reference image and matches state-of-the-art performance with two images on  MyVLM dataset. On  Yo'LLaVa dataset, we achieve comparable performance to Yo'LLaVa~\cite{nguyen2024yo} with only three images, even though the full set  includes up to 10 images for some objects.

\subsection{Number of Personalized Objects (\textit{P})}
\begin{figure}[h]
    \centering
    \includegraphics[width=0.6\linewidth]{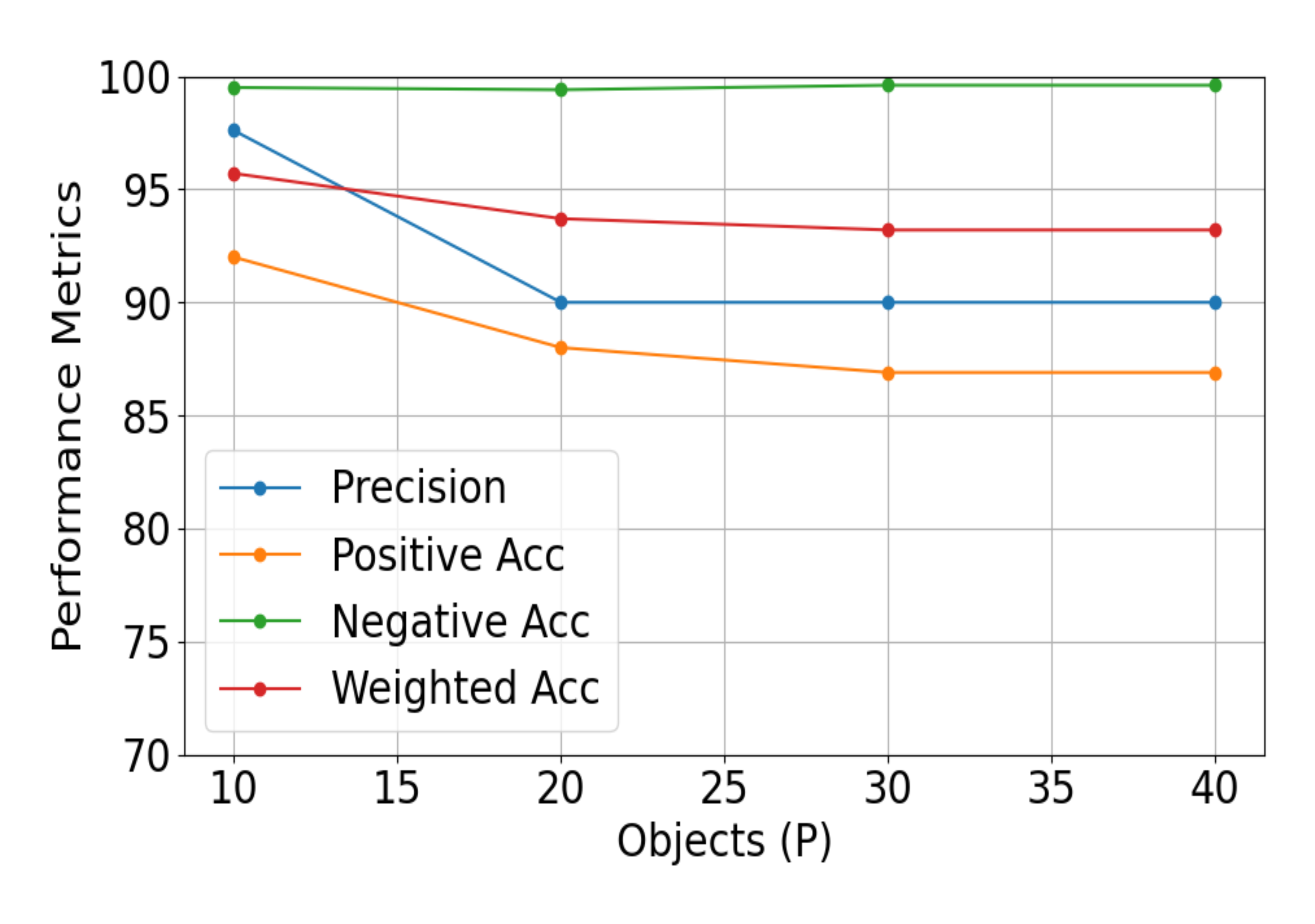}
    \caption{Ablation on P: \ours's performance when evaluated on a subset of 10 categories from Yo’LLaVA dataset while progressively increasing number of objects.} 
    \label{fig:number_obj}
\end{figure}

Another key consideration for any personalization method is its robustness to increasing number of personalized objects, particularly as more intra-category instances are introduced into the dataset. Figure~\ref{fig:number_obj} illustrates the performance of \ours\ on the first 10 objects from the \yollava\ dataset as the number of personalized objects increases incrementally from 10 to all 40 categories. While there is a slight performance drop at higher values of P, PeKit demonstrates overall stability and robustness.

\subsection{Threshold Selection}
In the main paper, we employ a fixed threshold of $\tau = 0.75$ across all experimental settings to determine whether a personalized object is present in an image.

\begin{figure*}[t]
    \centering
    \includegraphics[width=0.7\textwidth]{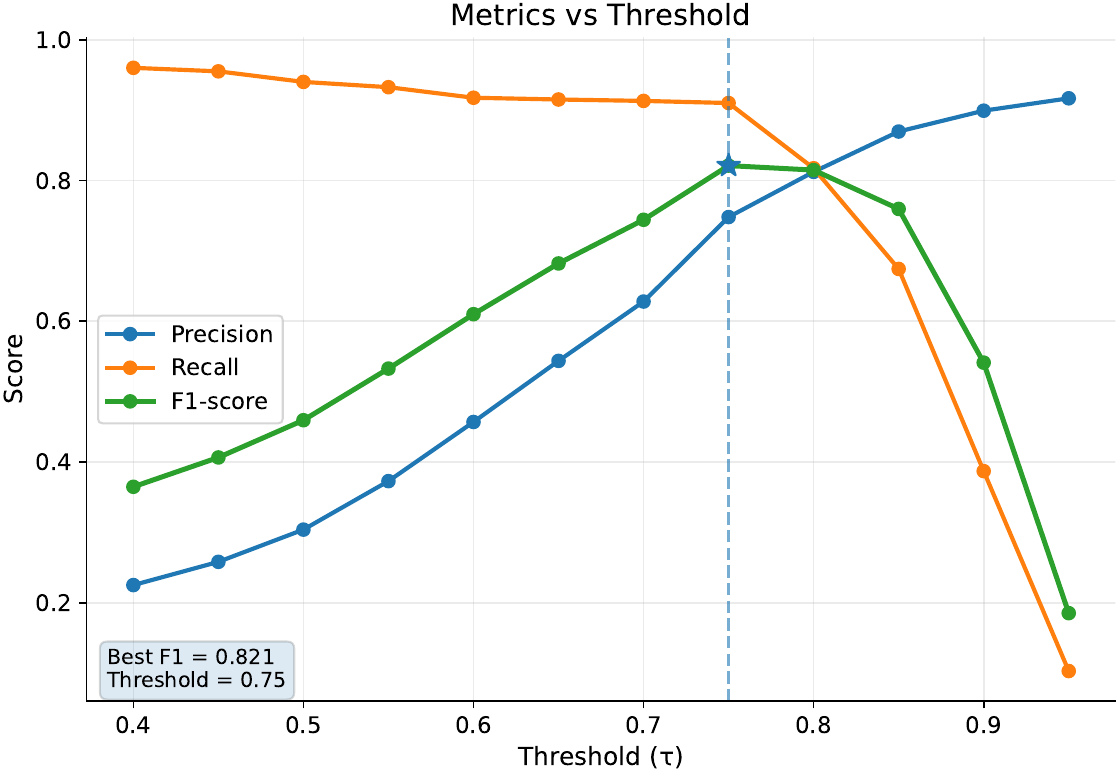}
    \caption{Precision, Recall and F1 score curve for different detection thresholds on \yollava dataset. The best performing threshold (0.75) in terms of F1 score is marked with a star.}
    \label{fig:PRF1}
\end{figure*}

\begin{figure*}[t]
    \centering
    \includegraphics[width=0.7\textwidth]{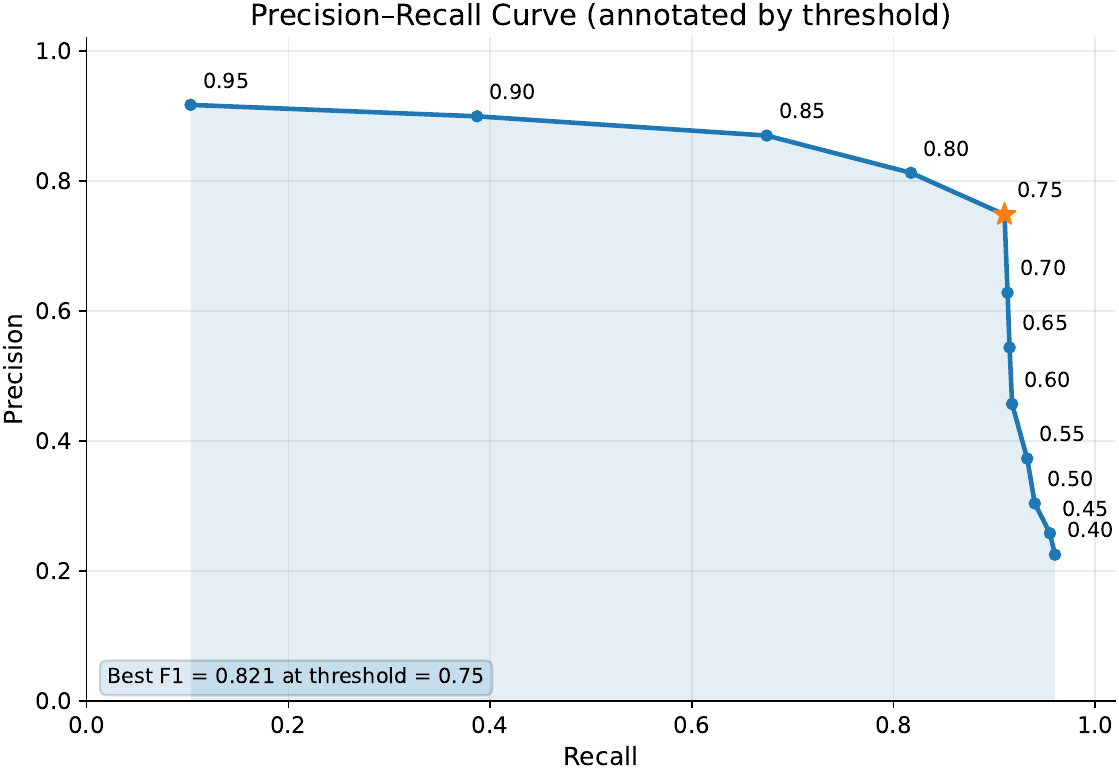}
    \caption{PR-curve on \yollava dataset. The threshold achieving the best F1 score is marked with a star.}
    \label{fig:thresh_PR}
\end{figure*}

Figures~\ref{fig:PRF1} and~\ref{fig:thresh_PR} illustrate the precision–recall trade-off for different values of $\tau$ on the \yollava{} dataset. As shown, $\tau = 0.75$ achieves the highest F1‑score, providing the best balance between precision and recall.

Lowering the threshold increases the number of detections but also raises the likelihood of false positives. Conversely, increasing the threshold improves precision at the cost of reduced recall, potentially causing some objects to be overlooked.

For instance, in a personalized search task such as the query “find my keys,” a high threshold may prevent the model from detecting the keys altogether, whereas a low threshold may return multiple key candidates, allowing the user to manually identify the correct one. In this study, we selected the threshold $0.75$ that best balances precision and recall on the \yollava dataset and applied it across all experiments, while noting that the threshold may be adjusted depending on the requirements of a specific application.

\label{sec:experiments}

\label{sec:experiments}

For completeness and flexibility, we further provide a straightforward method for tuning the thresholds on a per-object basis. To adjust the threshold for a given  
personalized object $i$, we first compute the distribution of distances for the reference images of the object, and then calculate the mean $\mu_i$ and standard deviation $\sigma_i$ of the distribution. The threshold is then set as $\tau_i = \mu_i - \sigma_i/2$. We note that by using this method, one can easily adjust the thresholds to be stricter or more loose by adding a scalar multiplier, as $\tau_i = \mu_i - \alpha (\sigma_i/2)$. Table~\ref{tab:supp_tuned} shows that this method can produce even higher accuracy.

 \begin{table}[h!]
 \centering
      \caption{Automatic object threshold selection.}
    \resizebox{0.7\linewidth}{!}{
\small
\begin{tabular}{|c|c|c|c|c|}
\hline
\multicolumn{5}{|c|}{\textbf{Yo' LLaVA Dataset}} \\
\hline
\multirow{2}{*}{Method/Metric} & \multicolumn{1}{c|}{Precision} & \multicolumn{3}{c|}{Recall} \\
 & & Positive & Negative & Weighted \\
\hline
Fixed $\tau = 0.75$ & \textbf{74.8} & \underline{91} & \textbf{98.7} & \underline{94.9}\\
\hline
Tuned $\tau = \mu - \sigma/2$  & \underline{66} & \textbf{92.5} & \underline{98.1} & \textbf{95.3}\\
\hline

\end{tabular}
}
      \label{tab:supp_tuned}
\end{table}

\subsection{Video-QA Object Naming}
Object names often offer important contextual clues about the subject in a query to the base LVLM. For example, when asked `What is Jack doing?', the model may infer that `Jack' refers to a male individual in the image. In such cases, the LVLM can often respond accurately without the need for personalization. To isolate the effect of personalization, we perform an ablation study on an open-ended video QA task, where all identified personalized objects are visually prompted using the generic label ENTITY (see Appendix~\ref{vprompting} and Fig.\ref{fig:supp_query} for prompt templates). Table~\ref{tab:vlm_variations} compares our method's performance when using object names or the term ENTITY to the base LVLM models.

 \begin{table}
 \small
 \centering
 \caption{Video-QA naming bias ablation. Our method mitigates naming bias, improving baseline VLM accuracy in ambiguous settings without explicit object names.}
\resizebox{0.8\linewidth}{!}{\small
\begin{tabular}{|c|c|c|c|c|}
\hline
\textbf{Model/Variation} & \textbf{Base} & \textbf{\ours} & \textbf{Base+ENTITY} & \textbf{\ours+ENTITY}\\
\hline
\textbf{LLaVA-OneVision} & 23.0 & \textbf{35.0} & 14.0 & \underline{19.9}\\
\hline
\textbf{InternVideoChat} & 56.6 & \textbf{61.3} & 45.8 & \underline{56.2}\\
\hline
\end{tabular}
}
\label{tab:vlm_variations}
\end{table}

\subsection{Time and Memory Requirements}
Table \ref{tab:supp_time_memory} details the memory usage and runtime of the various modules in our method for the \VLM personalization task. By default, we employ GroundingDINO (Base)~\cite{liu2023grounding} and SAM (Base)~\cite{kirillov2023segment} for extracting views from reference images and proposing objects during inference, DINOv2 (Large) features for retrieving personalized instances, and LLaVA 1.5 13B as our \VLM for generating answers. All experiments were conducted using an A5000 RTX GPU. As shown in the table, our plug-and-play modules are efficient, adding minimal overhead to the original LLaVA model.

Specifically, our approach introduces an overhead of 0.35 seconds per image (0.31 seconds from G-DINO) relative to LVLM's inference time. This overhead can be substantially reduced by employing a lightweight object proposal network such as YOLO-World~\cite{Cheng2024YOLOWorld}, which runs at 55 FPS compared to G-DINO’s 3 FPS. However, as reducing computational time is not the primary focus of this work, we leave further optimization in this direction for future research.

 \begin{table}[h!]
 \centering
       \caption{Time and memory requirement of \ours. View extraction is conducted only on the reference views corresponding to each personalized object category, whereas the remaining modules are executed for every image during inference.}
    \resizebox{0.7\linewidth}{!}{
\begin{tabular}{|c|c|c|c|}
\hline
\small
Module & Backbone & Time/Image (S) & Memory (GB) \\ \hline
View Extraction & G-DINO + SAM & 0.48 & 1.2  
 \\ 
 \hline
 Object Proposal & G-DINO & 0.31 & 0.87\\
Retrieval & DINOv2 & 0.04 & 1.8 \\ 
Reasoning & LLaVA 1.5 13B & 0.87 & 16.8\\
\hline
\end{tabular}
}
      \label{tab:supp_time_memory}
\end{table}




\section{This-Is-My-Img Benchmark Details}
The original This-Is-My dataset~\cite{yeh2023meta}, designed for video-level detection of personalized objects, comprises 104 training and 582 validation short video segments spanning 15 categories. However, some of the segments are no longer available for download. Consequently, one personalized object category, ‘Alex’s Piano’ has been removed from our proposed benchmark due to the unavailability of its training segments. 
\subsection{Single-concept Set}
\begin{figure}[!htbp]
    \centering
    \includegraphics[width=\textwidth]{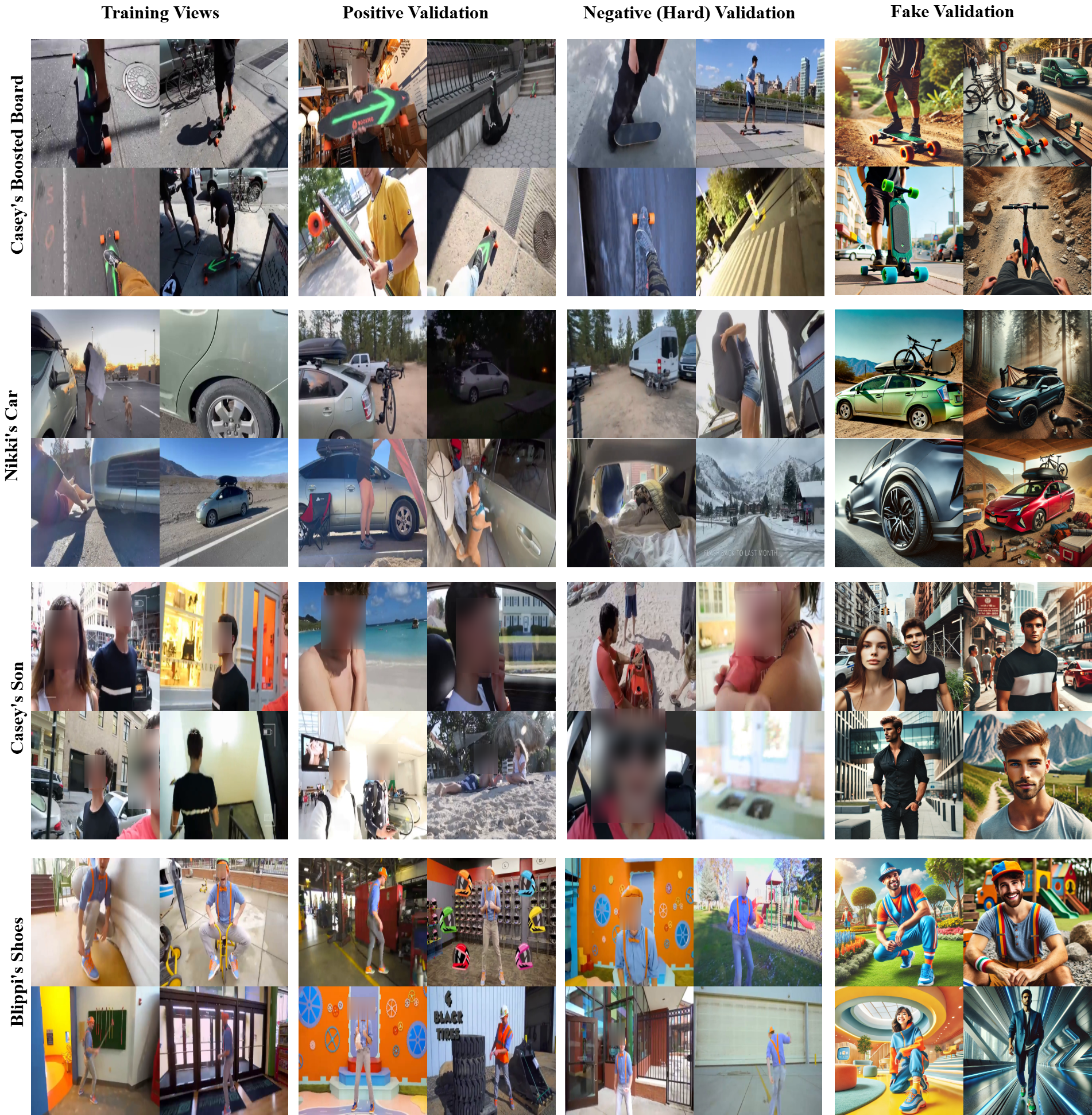}
    \caption{This-Is-My-Img single-concept benchmark. Our benchmark includes a wide range of concepts presented in realistic indoor and outdoor environments. Reference views can occasionally be sub-optimal, which increases the difficulty of the task. The positive validation set may contain false positives from within the same semantic category, allowing us to assess a model’s robustness to contextual similarities. The negative (hard) and Negative (fake) validation sets are crafted to challenge model resilience by imitating the appearance of personalized objects or their surrounding context. Additionally, the negative ( 'other') set—though not shown in this image—includes images of all other personalized objects for each concept, serving to evaluate the method's robustness to dataset bias.}
    \label{fig:supp_this_is_more}
\end{figure}

Table~\ref{tab:supp_this_is_my} presents further details about the personalized objects and the number of frames included in the single-concept validation splits of our benchmark. We sampled every 10th frame of each validation video segment and manually assigned frames to the respective splits. Note that three categories lack a \textbf{Negative (Hard)} set because the corresponding objects are present in every frame of their validation segments. The number of \textbf{Positive}, \textbf{Negative (Hard)}, and \textbf{Negative (Other)} frames varies between categories due to differences in segment length and the number of segments per object. In contrast, the \textbf{Fake} set was standardized, with 10 validation frames generated for each class. Figure~\ref{fig:supp_this_is_more} offers additional visual examples from our benchmark.

The single-concept validation also includes a VQA set comprising 70 images (5 per category). Challenging frames were manually selected, and initial question–answer pairs were generated using the GPT-4O model. These pairs were then manually refined to create a high-quality evaluation set. In line with the \yollava framework, answers for this set are presented in an A/B multiple-choice format.

 \begin{table}[h!]
 \centering
 \caption{Single-concept categories and number of frames of \thisismy benchmark. The validation frames in the dataset are organized into three subsets for each category: frames where the object is visible (\textit{Positive}), frames from the same video segments where the object is absent (\textit{Hard}), and positive frames belonging to other object categories (\textit{Other}). In addition, the benchmark provides a GPT-generated set of 10 images per category (\textit{Fake}).}
    \resizebox{0.7\linewidth}{!}{\begin{tabular}{|c|c|c|c|}
\hline
\small
Category & Positive& Negative (Hard)& Negative (Other)\\ \hline
Alex's Bag & 161 & 43 & 2096 \\
Alex's Hat & 120 & 0 & 2137 \\ 
Blippi's Shoes & 339 & 79 & 1918  \\
Casey's Boosted Board & 35 & 118 & 2222  \\
Casey's Friend Marlan & 24 & 4 & 2233\\
Casey's Son & 46 & 15 & 2211 \\
Gab's Puppy Lili & 56 & 12 & 2201 \\
Nikki's Camper Bag & 229 & 76  & 2028 \\
Nikki's Car & 651 & 184 & 1606\\
Reynard's Keyboard & 162 & 29 & 2095 \\
Reynard's Work Chair & 188 & 59 & 2069 \\
Sherry's Road Bike & 95 & 12 & 2162 \\
Zak's Dog Coffee & 26 & 0 & 2231 \\
Zak's Dog Kona & 125 & 0 & 2132 \\
\hline
Sum & 2257 & 631 & 29341\\
\hline
\end{tabular}
}
      \label{tab:supp_this_is_my}
\end{table}
\subsection{Multi-concept Set}
Table~\ref{tab:supp_this_is_my_multi} outlines the category pairs used in our multi-concept validation benchmark. This benchmark comprises 55 images, with each category pair represented by 5 manually selected positive examples sourced from the validation frames. We extended the original \thisismy dataset categories (Table~\ref{tab:supp_this_is_my}) by incorporating new personalized categories: Alex, Blippi, Casey, Gab, Nikki, Sherry, and Zak. For each of these added categories, we selected 5 reference views from the training segments. Each image in the benchmark is accompanied by an open-ended question–answer pair, collaboratively crafted with GPT-4O. This open-ended visual question answering (VQA) format raises the task’s difficulty and reduces the likelihood that the underlying LVLM model can succeed through answer-choice elimination alone.
 \begin{table}[h!]
 \centering
   \caption{Multi-concept categories on \thisismy benchmark. Each category-pair comes with 5 positive frames and 50 negative frames from other category pairs.}
\resizebox{0.7\linewidth}{!}{\scriptsize
\begin{tabular}{|p{4cm}|p{4cm}|}
\hline
\multicolumn{2}{|c|}{\textbf{Category-pair}} \\
\hline
Alex - Alex's Bag & Nikki - Nikki's Car \\ 
Alex - Alex's Hat & Nikki - Nikki's Camper Bag \\
Blippi - Blippi's Shoes & Sherry - Sherry's Road Bike \\
Casey - Casey's Boosted Board & Zak - Zak's Dog Coffee \\
Casey - Casey's Son & Zak - Zak's Dog Kona \\
Gab - Gab's Puppy Lili & \\
\hline
\end{tabular}
}
      \label{tab:supp_this_is_my_multi}
\end{table}
\subsection{Video-QA Set}
As shown in the main paper, \ours can be extended to video personalization by applying the method to sampled frames. To evaluate this, we used validation video segments (typically under 15 seconds) and curated a VQA dataset with 267 high-quality question–answer pairs for the single-concept categories in Table~\ref{tab:supp_this_is_my}. For each video segment, we employed the LLaVA-OneVision-Qwen2-7B model with a 1-in-16 frame sampling rate to generate an initial set of 1,380 question–answer candidates. These were then filtered for duplicates using GPT-3.5 Turbo, reducing the set to 618 pairs. Finally, we manually refined the results to produce a curated set of 267 high-quality QA pairs.

\section{Prompt Templates}\label{vprompting}
In this section, we present the specific query format used across different parts of our experiments.
\subsection{Visual prompting}
As mentioned in the main paper (Sec 3.4), we employ visual prompting to indicate the personalized object’s location in the image. Next we query our \VLM to personalize its answer given the instance's name and context.  Our prompt to the \VLM for various tasks follows this general structure:

\begin{center}
In this image (video), the entity enclosed in a `\textbf{COLOR}' box is called `\textbf{NAME}'. 

Without mentioning the bounding box and its color, {`\textbf{TASK}'}.

[Optional] Give more details using the information from {`\textbf{CONTEXT}'}.

\end{center}

The \textbf{COLOR} placeholder indicates the color of the bounding box overlaid on the image, the \textbf{NAME} placeholder specifies the instance name, and the \textbf{TASK} placeholder contains the task-specific query. Optionally, the {\textbf{CONTEXT}} placeholder can contain the prior knowledge about the personalized object retrieved from the memory module.

Note that when multiple objects are detected in one image, the query’s grammatical structure changes to a plural format, and the \textbf{COLOR} and \textbf{NAME} placeholders will contain multiple values separated by commas. The \textbf{CONTEXT} placeholder for each object is included in angle brackets ($<>$) and contains the \textbf{NAME} for the corresponding instance. 

For the experiments in the paper the \textbf{TASK} placeholder can be any of the following prompts:

\begin{center}
\textbf{Personalized Captioning}: Describe what is `\textbf{NAME}' doing. Describe the image too.

\textbf{VQA}: Answer the following question about `\textbf{NAME}': `\textbf{QUESTION}'.
\end{center}

\begin{figure*}[h]
    \centering
    \includegraphics[width=\textwidth]{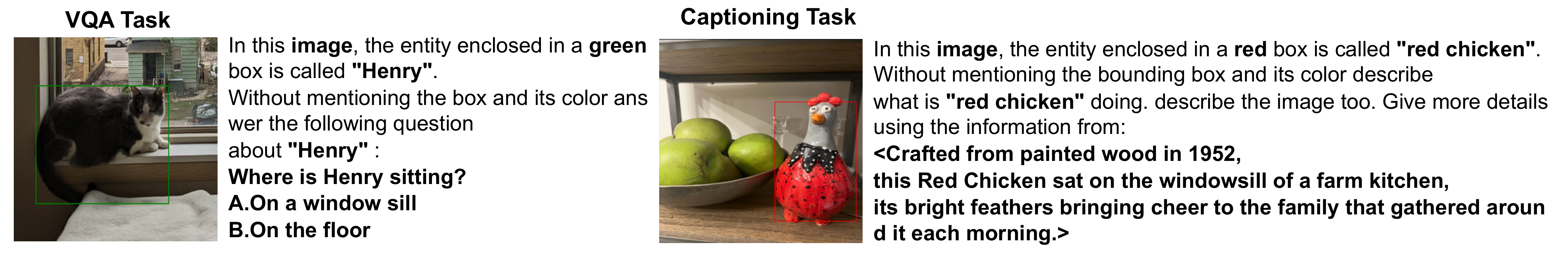}
    \caption{Prompt format. Personalized VQA and captioning on Yo'LLaVA (Left) and MyVLM (Right) datasets. The context used for the `red chicken' is imaginary and generated by ChatGPT.}
    \label{fig:supp_query}
\end{figure*}

Figure~\ref{fig:supp_query} illustrates an example of our full prompt for each one of the tasks.

\subsection{Open-ended VQA Validation}
To evaluate the accuracy of model predictions in an open-ended VQA task, we adopt the evaluation pipeline proposed by~\cite{maaz2023video}. In particular, we utilize the following prompt template to query a GPT-3.5-Turbo model for assessing the semantic alignment between predicted answers and ground truth responses for each question in the VQA dataset.

\begin{center}
`You are an intelligent chatbot designed for evaluating the correctness of generative outputs for question-answer pairs. Your task is to compare the predicted answer with the correct answer and determine if they match meaningfully. 

Here's how you can accomplish the task:
INSTRUCTIONS: 
\begin{itemize}
\item Focus on the meaningful match between the predicted answer and the correct answer. 
\item Consider synonyms or paraphrases as valid matches.
\item Evaluate the correctness of the prediction compared to the answer.
\end{itemize}
Please evaluate the following video-based question-answer pair: 

Question: {\textbf{QUESTION}} 

Correct Answer: {\textbf{ANSWER}} 

Predicted Answer: \textbf{PREDICTION} 

Provide your evaluation only as a yes/no answer. Please generate the response in the form of a Python dictionary string with key `pred', where value of `pred' is a string of `yes' or `no'.

DO NOT PROVIDE ANY OTHER OUTPUT TEXT OR EXPLANATION. Only provide the Python dictionary string. For example, your response should look like this: {`pred': `yes'}.'
\end{center}

We calculate the VQA accuracy by dividing the number of times the GPT model, using the specified prompt template, responds with `yes' by the total number of question-answer pairs in the set.

\section{Qualitative Results}

\begin{figure*}[t]
    \centering
    \includegraphics[width=\textwidth]{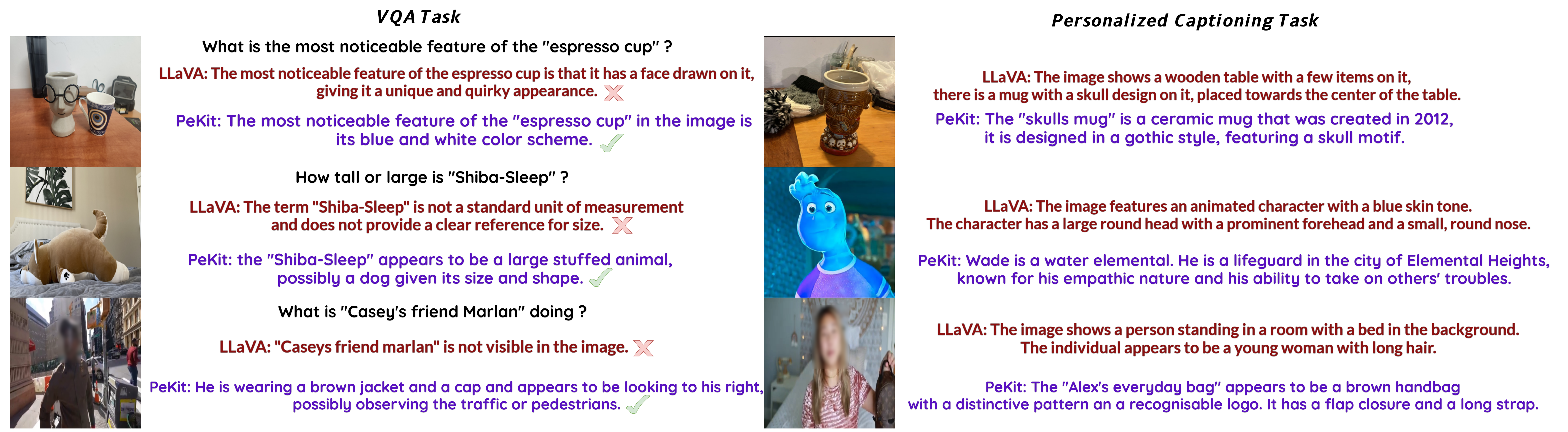}
    \caption{Qualitative comparison to LLaVA. Right: Our method detects personalized objects and integrates provided context (for qualitative comparison) in caption generation. Left: While the original model struggles with specific questions about named objects, our method easily identifies the referred object.}
    \label{fig:qualitative_llava}
\end{figure*}

\begin{figure*}[!htbp]
    \centering
    \includegraphics[width=\textwidth]{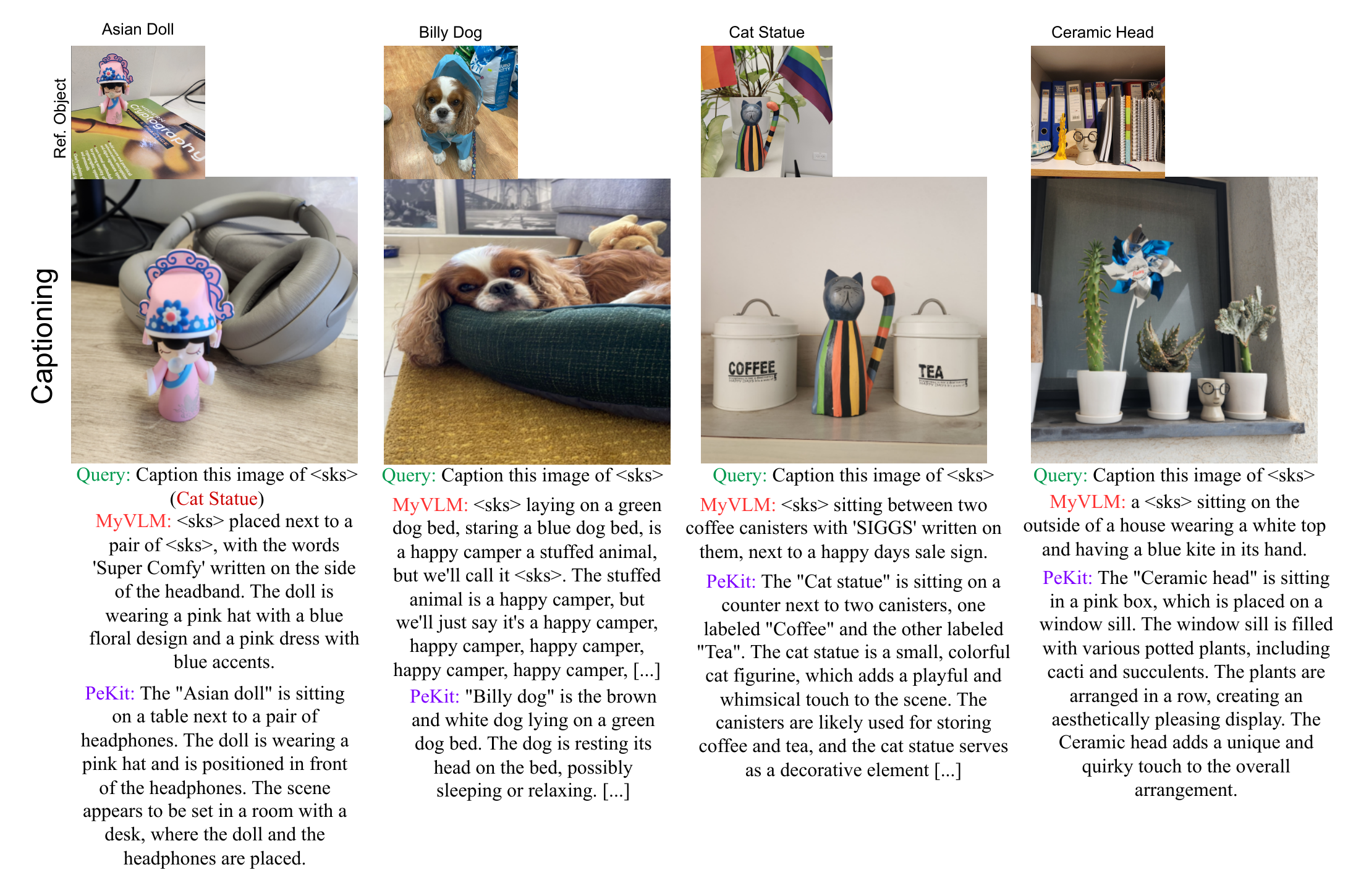}
    \caption{Qualitative comparison to MyVLM. MyVLM often misidentifies personalized objects because of its low precision. In the leftmost figure, when prompted to caption an image containing a `Cat Statue'—which is actually absent—MyVLM incorrectly labels the `Asian doll' and the headset as the `Cat Statue' instead of rejecting the query. Additionally, MyVLM training interferes with the original captioning capabilities of the LVLM, leading to hallucinations, short captions, and sometimes incomprehensible text. For each image, `Query' depicts MyVLM's sytem prompt where the concept identifier $<$sks$>$ is replaced with the personalized object's name. PeKit employs its own prompt template described in Appendix C.1.}
    \label{fig:my_comparison}
\end{figure*}

\begin{figure*}[!htbp]
    \centering
    \includegraphics[width=\linewidth]{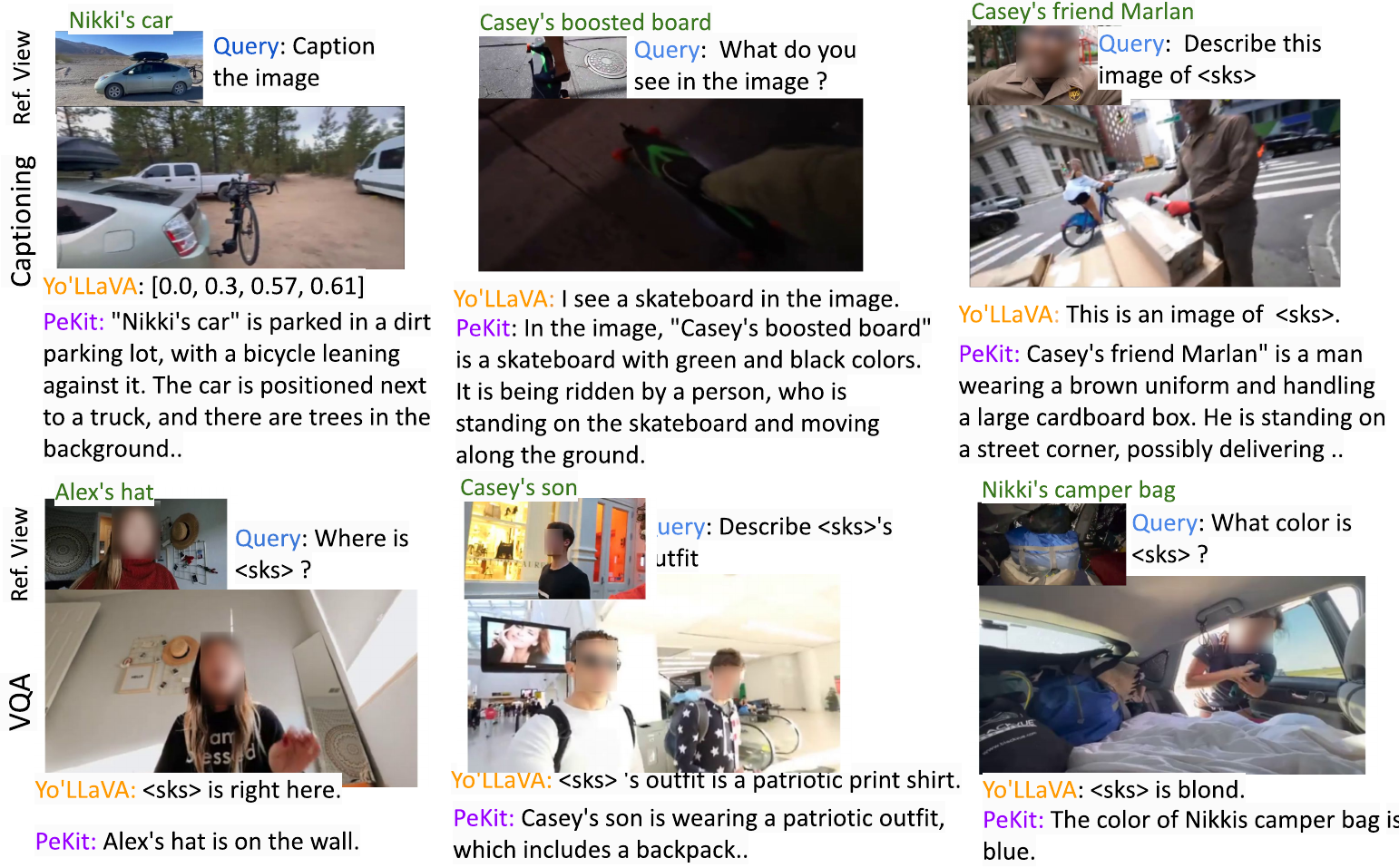}
        \caption{Qualitative comparison to Yo'LLaVA. Yo’LLaVA’s prompt template requires specifying the personalized object's identifier in the query (first row), limiting generalization since users must already know which objects are in the image. Using image-level embeddings can also cause confusion between similar objects (e.g., Alex vs. Alex’s bag). Adjusting the LLM’s head weights further harms captioning quality. \ours achieves better captioning quality without any training. In each example, the `Query' shows \yollava's prompt with the concept identifier replaced by the object's name and an added system prompt. \ours uses a different template, detailed in the Appendix C1.}
    \label{fig:qualitative_yollava}
\end{figure*}

\subsection{Qualitative Comparison to LLaVA}
Figure~\ref{fig:qualitative_llava} presents examples of \ours model compared to the base LLaVA~\cite{liu2024visual} on VQA and personalized captioning tasks using images from all three benchmarks discussed in the main paper. When LLaVA does not recognize an object from the given name in the query, it makes guesses, leading to hallucinations or incorrect statements. In contrast, \ours accurately identifies objects and uses in-context information to guide LLaVA in answering questions or providing details about the image, effectively incorporating in-context information and object appearance. While more advanced prompting or personalized response examples could enhance \ours, we opted for simplicity and standard design, leaving such improvements for future work.

\subsection{Qualitative Comparison to MyVLM \cite{alaluf2024myvlm}}
Figure~\ref{fig:my_comparison} compares PeKit with MyVLM on images from the MyVLM dataset. We used the original checkpoints and code provided by the authors of MyVLM to generate the results. Checkpoints for the VQA task were not provided.

MyVLM shares the same limitation as Yo'LLaVA, functioning exclusively when the concept identifier is included in the query. Additionally, MyVLM exhibits low precision in detecting personalized objects, as shown in the main paper for the \thisismy benchmark. This can lead to misidentifying objects as personalized ones. The leftmost image in Figure~\ref{fig:my_comparison} demonstrates this limitation. MyVLM is asked to provide a caption for the target concept ‘Cat Statue’ while the provided image includes another personalized object, ‘Asian Doll.’ As shown, MyVLM incorrectly identifies the ‘Asian Doll’ as the ‘Cat Statue’ and generates an incorrect personalized caption. Our method addresses this issue by first detecting the correct personalized object(s) and then generating a caption based on the prompt template provided in Appendix C.1.

Furthermore, MyVLM’s training appears to degrade the original LVLM’s captioning capabilities, leading to short captions with hallucinations and sometimes incomprehensible text, leading to a low CLIPScore as demonstrated in the main paper.

\subsection{Qualitative Comparison to \yollava \cite{nguyen2024yo}}
Figure~\ref{fig:qualitative_yollava} compares \ours to \yollava for VQA and personalized captioning tasks on \thisismy benchmark. As seen on the first row, \yollava's prompt template requires the query to include the target object’s concept identifier, making it unsuitable for general captioning tasks and tailored for Visual Question Answering. Besides, since \yollava operates on image-level embeddings of reference views, it needs clutter-free object-centered reference views of the personalized objects. As seen on the second row, performance can decline if the personalized object is not in the foreground or if there are other objects/people interacting with the personalized object in the reference views. Besides, fine-tuning the last layer of the language model reduces the LVLM’s captioning capabilities for \yollava. \\

\subsection{Real-world Demonstration}\label{sec:real_world}

\begin{figure*}[h]
    \centering
\includegraphics[width=\textwidth]{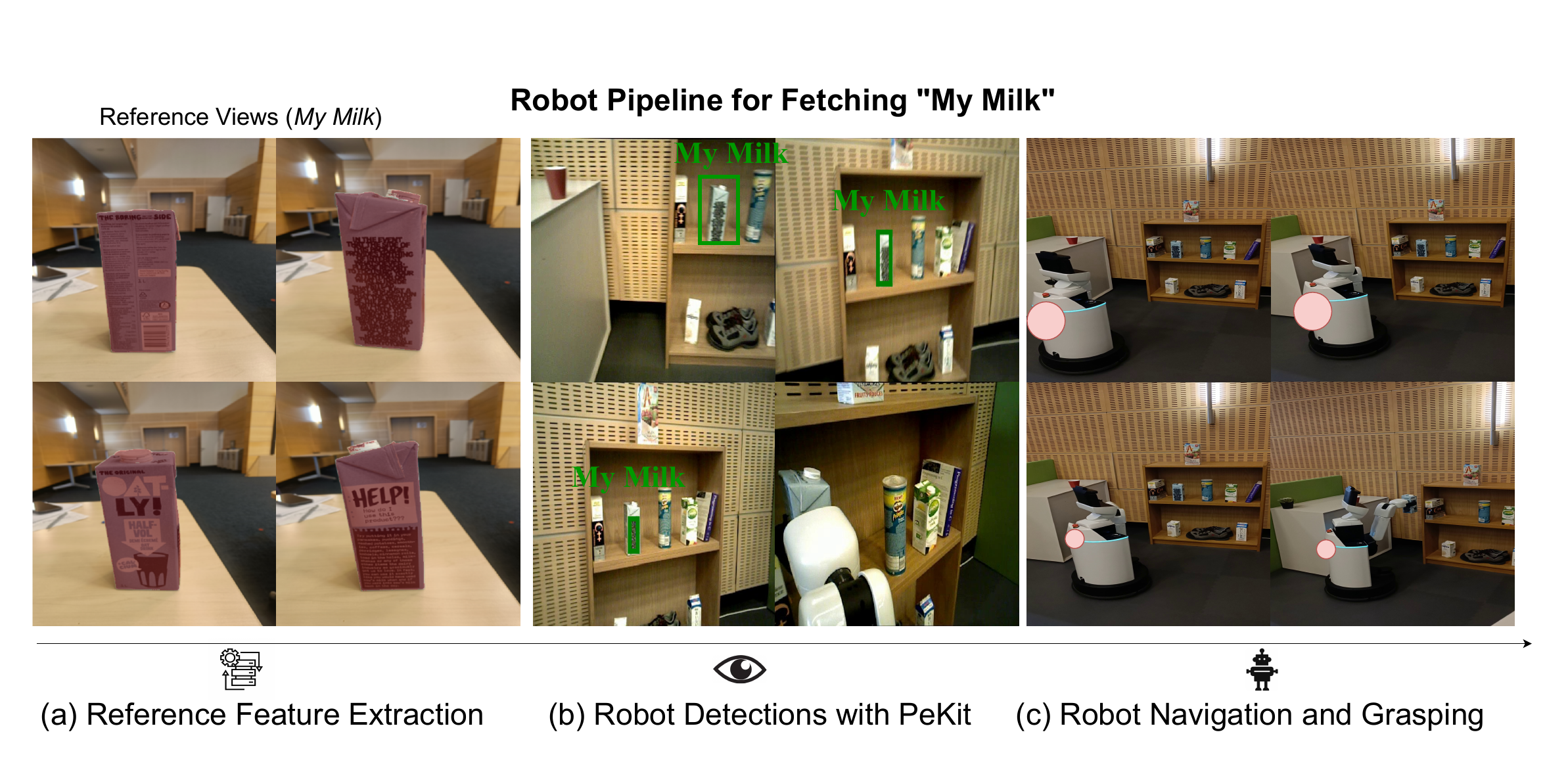}
    \caption{Real-world demonstration. \ours can be incorporated into a mobile manipulation robot to perform personalized object search and fetching.}
    \label{fig:hsr_experiment}
\end{figure*}

To evaluate \ours in real-world conditions, we deploy it on a mobile manipulation robot for a “search and fetch” task (Figure~\ref{fig:hsr_experiment}). The robot autonomously explores an unfamiliar environment to locate and grasp a specific object, in this case a carton milk box named `My Milk'.

The robot, equipped with a wheeled base and a single-arm manipulator, combines mobility and dexterity for effective navigation and interaction. Exploration is guided by MORE \cite{more25mohammadi}, a pipeline that uses scene graphs and large language models (LLMs) to plan actions from a 3D panoptic map generated by OpenVox \cite{deng2025openvox}. We enhance OpenVox with occupancy mapping and integrate \ours to refine object labels for accurate identification.

Before deployment, \ours's RAG memory is initialized using a small set of target object images processed through our view extraction pipeline. Refined labels and the panoptic map are shared with MORE via ROS.
To grasp the object, the robot navigates to its location, captures an RGB-D image, extracts the object mask using projected panoptic data refined by SAM \cite{kirillov2023segment}, and estimates a grasping pose via a rule-based method. This enables successful retrieval of specific items—e.g., a chosen milk carton from a shelf with similar packages.

\end{appendices}

\end{document}